\PassOptionsToPackage{numbers, compress}{natbib}
\PassOptionsToPackage{prologue,dvipsnames}{xcolor}

\documentclass{article}

\usepackage[preprint]{neurips_2025}

\usepackage[utf8]{inputenc} 
\usepackage[T1]{fontenc}    
\usepackage{hyperref}       
\usepackage{url}            
\usepackage{booktabs}       
\usepackage{amsfonts}       
\usepackage{nicefrac}       
\usepackage{microtype}      
\usepackage{xcolor}         
\usepackage{wrapfig}

\DeclareUnicodeCharacter{FF0C}{,}
\usepackage{graphicx}
\usepackage{subfigure}
\usepackage{booktabs} 

\makeatletter
\def\rulefill{\leavevmode\leaders\hrule height .7ex width 1ex depth -0.6ex\hfill\kern\z@}
\makeatother

\usepackage{amsmath}
\usepackage{amssymb}
\usepackage{mathtools}
\usepackage{amsthm}

\usepackage[capitalize,noabbrev]{cleveref}

\theoremstyle{plain}
\newtheorem{theorem}{Theorem}[section]

\theoremstyle{definition}
\newtheorem{definition}[theorem]{Definition}
\newtheorem{assumption}[theorem]{Assumption}
\theoremstyle{remark}

\usepackage{pifont}
\newcommand{\cmark}{\ding{51}}%
\newcommand{\xmark}{\ding{55}}%

\usepackage[textsize=tiny]{todonotes}

\usepackage{notation}
\usepackage{comment}
\usepackage{array}
\usepackage{multirow}
\usepackage{makecell}
\usepackage{siunitx}
\usepackage{tabularray}
\usepackage{enumitem}
\usepackage{colortbl}
\usepackage{tikz}
\usetikzlibrary{automata, positioning, arrows}
\usetikzlibrary{arrows.meta}
\usepackage{fancybox}
\usepackage{subcaption}
\usepackage{adjustbox}
\usepackage{caption}

\UseTblrLibrary{booktabs, siunitx} 

\title{Learning Model Successors}

\author{Yingshan Chang \\
  Carnegie Mellon University\\
  Pittsburgh, PA 15213 \\
  \texttt{yingshac@andrew.cmu.edu} \\
  \And
  Yonatan Bisk \\
  Carnegie Mellon University\\
  Pittsburgh, PA 15213 \\
  \texttt{ybisk@andrew.cmu.edu}
}

\begin{document}

\maketitle

\begin{abstract}
The notion of generalization has moved away from the classical one defined in statistical learning theory towards an emphasis on out-of-domain generalization (OODG). There has been a growing focus on generalization from easy to hard, where a progression of difficulty implicitly governs the direction of domain shifts. 
This emerging regime has appeared in the literature under different names, such as length/logical/algorithmic extrapolation, but a formal definition is lacking. We argue that the unifying theme is induction --- based on finite samples observed in training, a learner should infer an inductive principle that applies in an unbounded manner. 
This work formalizes the notion of \textit{inductive generalization} along a \textit{difficulty progression} and argues that our path ahead lies in transforming the learning paradigm. We attempt to make inroads by proposing a novel learning paradigm, \textit{Inductive Learning}, which involves a central concept called \textit{model successors}. We outline practical steps to adapt well-established techniques towards learning model successors. This work calls for restructuring of the research discussion around induction and generalization from fragmented task-centric communities to a more unified effort, focused on universal properties of learning and computation. 

\vspace{-12pt}
\end{abstract}

\section{Introduction}
\label{sec:intro}

Neural sequence modeling with current learning paradigms often runs into problems that manifest as length-generalization failures \cite{turingprogramsscratchpad, lengen_arithmetic, PEimpact-on-lengen, n_rconsistency, zhou2024lengen}. This paper clarifies that one root cause of this bottleneck is the inability to extrapolate along a \textit{difficulty progression}, wherein the true difficulty indicator is not necessarily the input length, but the location of a testing instance along the progression. 

Take counting as an example -- the task of constructing a map between set sizes and integers. While the unseen vocabulary and unseen position embeddings incurred by a longer testing input can be addressed by auxiliary tasks and augmented position embeddings, respectively, unseen cardinalities resist any easy fix. \cite{yingshancounting} reveals that the failure to generalize to greater cardinality persists whenever the neural network architecture cannot express the desired inductive bias. 

Consider recognizing the $dyck_1$ language as another example -- the task of recognizing balanced brackets. Here, the true difficulty indicator is the nesting depth. Numerous studies revealed that prevailing neural networks cannot generalize to greater nesting depth unseen during training \cite{bounded_hierarchical_languages, learn_to_solve_recursively}. In particular, finite-precision RNNs cannot recognize $dyck_1$ of arbitrary depth because their computational power can be characterized by finite-state automata \cite{computational_power_rnns} whereas recognizing $dyck_1$ requires a computational power equivalent to pushdown automata \cite{merrill-sequentialNN_automata}. Transformers cannot even fit the training set of recognizing $dyck_1$ \cite{transformer_formal_language, hahn_limits_of_SA} without special techniques \cite{SA+} to ease learning.

Failing to generalize along a difficulty progression is not resulted from a limitation in data, model size, or inference time computation, but from limitations of a learning paradigm that does not allow for capturing high-order patterns. App.~\ref{app:motivating} provides three experiments that illustrate key problems and better motivate this paper. Despite the struggles faced by machines, humans find it trivial. For example, children first learn to count one, two, three, or four objects as if they were separate instances \cite{cogsci_sarnecka, wynn1992children}. Then they transition to realize that there are infinitely many integers and thus counting can proceed to infinity \cite{carey_origin}. The cognitive science literature characterizes this sharp transition as an inductive leap, where an inductive principle is inferred \footnote{
Informally, the inductive principle of counting states that adding one object to a set increases the size by one \citet{Rips_boot_to_bootstrap, margolis_learn_numbers}.
}. 

To capture this inductive principle, machines must move beyond pattern-finding in data to pattern-finding in models. This demands learning at multiple levels of abstraction and leaping from one hypothesis class to another --- a capacity absent in current practices. Bridging this gap requires changing the learning paradigm. This paper initiates such a paradigmatic shift by formalizing \textit{Inductive Learning}, where the hypothesis space at one level becomes the data space at the next level. We will differentiate these two levels via “base-learner” versus “inductive-learner”. The base-learner captures regularities in data and produces models. The inductive-learner captures regularities in models and produces what we call the “model successor”. Thus, the essence of Inductive Learning is \textit{learning model successors}. App.~\ref{app:motivating} demonstrates an example realization of learning model successors that achieves generalization to greater nesting depths on recognizing the $dyck_1$ language.

Our contributions lie in both (a) formalization of a novel learning paradigm, and (b) unification of the scientific language used for reasoning about learning paradigms. To achieve (a), we first propose a principled notion of difficulty progression \S~\ref{sec:difficulty_progression}. Then we formalize the framework of inductive learning \S~\ref{sec:DGR}\S~\ref{sec:inductive_learnability}, which accommodates research on a shared theme of ``easy-to-hard extrapolation" in a discrete input space. Finally, we outline practical steps toward learning model successors \S~\ref{sec:roadmap}, guiding the navigation of an interdisciplinary research landscape. To achieve (b) we upgrade the notation inherited from the rich learning theory literature \S~\ref{sec:notation}, bringing clarity to the notions of expressivity, learnability, and generalizability as distinct questions. We provide a taxonomy of learning paradigms \S~\ref{sec:learning_paradigms}, in which their essential differences can be articulated in light of our notation.

\section{Notation}
\label{sec:notation}

\newcommand{\support}[0]{$\cS$ }
\newcommand{\supportk}[0]{$\cS_k$ }

We follow notations established in learning theory \cite{ML-theory, Vapnik_statsLT} to describe probabilities, samples and hypotheses. We follow notations established in computational complexity theory \cite{universal_solomonoff, Hutter_universalAI, intro_to_Kolmogorov, AR-learnability} to describe discrete data in terms of strings.

\paragraph{Data, Distributions and Domains}
A data sample consists of input $x$ and output $y$ generated by $\mu$, written as $(x, y) \sim \PP_{\mu}$. Without loss of generality, suppose $x, y$ are strings (sequences) drawn from a unified alphabet (vocabulary) $\Sigma' = \Sigma_x \cup \Sigma_y$. Let `\_' be a novel character $\notin \Sigma'$. Then, let $\Sigma = \{\text{`\_'}\}\cup \Sigma'$. Hence, each data sample $(x, y)$ corresponds to a concatenated string $x\_y$. 
Denote the support by \support, which is the set of all strings with non-zero probability: \support$= \{a\ |\ a = x\_y, \PP_{\mu}(x, y)>0\}$. 
Denote a sample of size $n$ by $d^n \triangleq \{(x_i, y_i)\}_{i=1}^n$. Let $\cD^n$ be the set of all size-$n$ samples: $\cD^n = \big\{d^n\ |\ (x_i, y_i) \sim \PP_{\mu}\big\}$, and $\cD$ be the set of all possible samples regardless of sample size: $\cD = \{\cD^n\ |\ n \in \NN\}$. We call such a $\cD$ a \textit{domain}.
Since an input-output pair $(x, y)$, a string $x\_y$ and a sample $d$ all follow distributions determined by $\mu$, with a slight abuse of notation, we can write $x\_y \sim \PP_{\mu}, d \sim \PP_{\mu}, d^n \sim \PP_{\mu}$\footnote{
We may drop the superscript $n$ when sample complexity is not of immediate relevance to the discussion.
}.
When there are $k$ ordered domains, $\cD_1, ..., \cD_k$, each $\cD_i$ having probability $\PP_{\mu_i}$ and support $\cS_i$, denote $\cD_1 \times ... \times \cD_k$ as $\cD_{\leq k}$. Similarly, we can obtain samples $d_{\leq k} = (d_1, d_2, ..., d_k)$\footnote{
We use ``$()$" instead of ``$\{\}$" to emphasize that $d_{\leq k}$ is \textit{ordered}.
}. It is easy to see $d_{\leq k} \in \cD_{\leq k}$.

\begin{table*}[t]
\setlength{\belowrulesep}{3pt}
\setlength{\aboverulesep}{3pt}
\centering
\scriptsize
\begin{tabular}{@{}lll@{}}
\textbf{Terminology} & \textbf{Text Statement} & \textbf{Formal Statement} \\
\toprule
(a) Expressivity & $\exists \text{Inv}$ across $\cD_1, ..., \cD_k$ & $ |\bigcap_{j \leq k} \cH_j^{\mathrm{Lr}}| > 0$ \\
(b) Expressivity & $\exists \text{Inv}$ across $\cD_1, ..., \cD_k$ and in unseen domain $\cD_m$ & $ |\bigcap_{j \leq k\ or\ j=m} \cH_j^{\mathrm{Lr}}| > 0, m>k$ \\
(c) Learnability & Provable learning of invariance-capturing hypotheses & w.p. $1-\delta$, $L(d_{\leq k}) \in  \bigcap_{j \leq k} \cH_j \subseteq \bigcap_{j \leq k} \cH_j^{\mathrm{Lr}}$ \\
(d) Generalizability & \makecell[l]{Provable learning of invariance-capturing hypotheses.\\ \text{Inv} also hold in unseen domain $\cD_m$} & \makecell[l]{w.p. $1-\delta$, $L(d_{\leq k}) \in \big(\bigcap_{j \leq k} \cH_j \big) \cap \cH_m $\\ \qquad\qquad\qquad$ \subseteq \bigcap_{j \leq k\ or\ j=m} \cH_j^{\mathrm{Lr}}, m>k$} \\
\bottomrule
\end{tabular}
\caption{Our notation builds consensus on formally stating expressivity, learnability and generalization. When multiple domains are involved, \textbf{Invariance} ($\mathrm{Inv}$) proves central to all statements. We use shorthands ``w.p." for ``with probability" and ``$L$" for ``learner".}
\label{tab:ELG_Inv}
\end{table*}

\paragraph{Expressible, Low-risk, and Feasible Hypotheses}
\label{sec:Ex_Lr_Fe_Hypotheses}

$h$ is a hypothesis that belongs to a hypothesis space $\cH$. $h^*$ is the optimal hypothesis with respect to some task and performance measure. $\hat{h}^*$ is a close approximation to the optimal hypothesis, which could be the output of a reasonably good learner $L$ given some training set $d$, i.e. $L(d) = \hat{h}^*$. 

Existing learning frameworks across multiple domains generally assume one fixed hypothesis class \cite{lifelongML, manyPAC}. Thus, we take some time to better motivate the need for differentiating hypotheses in the sense that learner and data together identify different subsets of \textit{feasible hypotheses}. 

\begin{wrapfigure}[15]{r}{0.35\linewidth}
\centering
  \vspace{-12pt}
  \includegraphics[width=\linewidth]{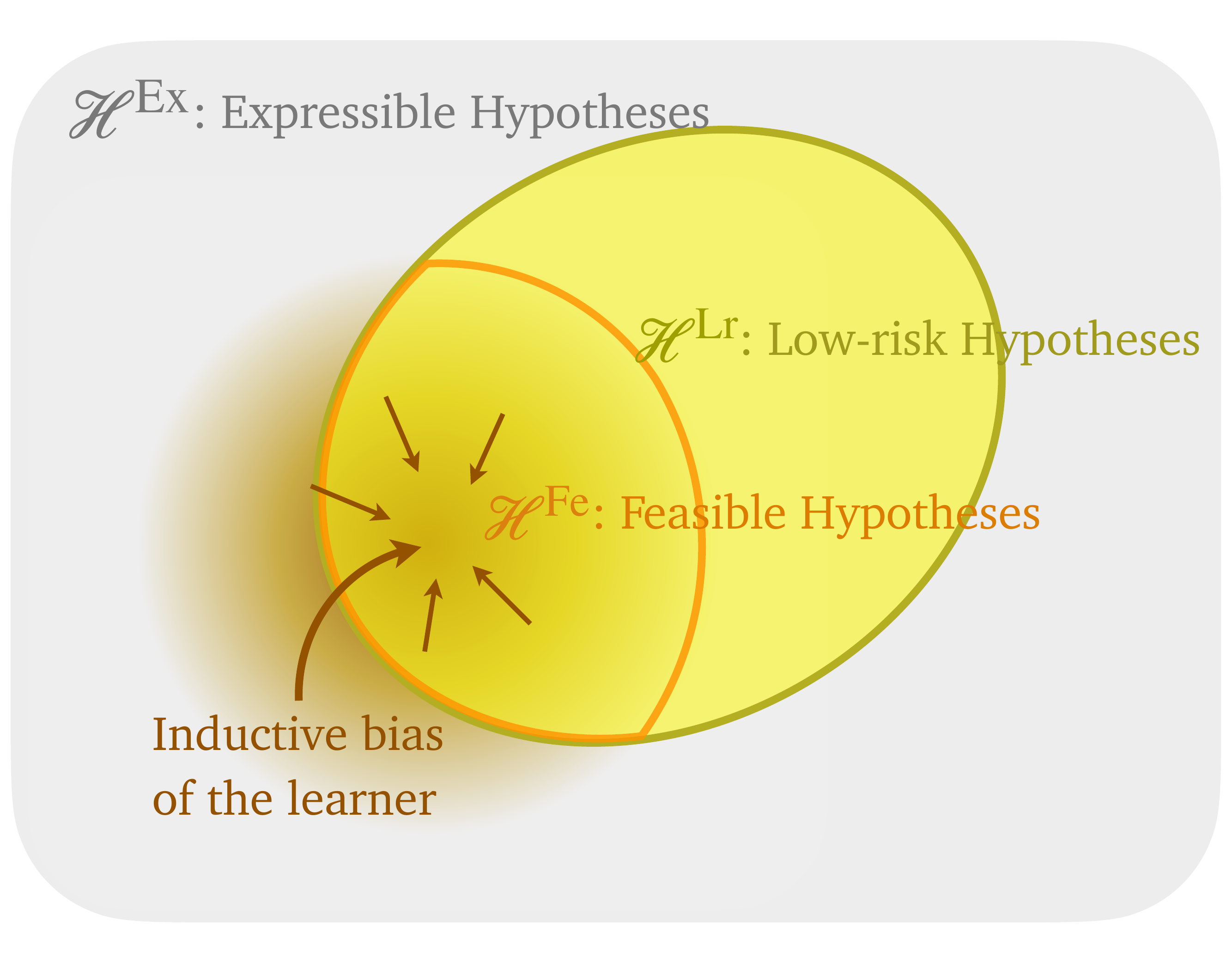}
  \vspace{-15pt}
  \caption{Hypotheses that are a priori preferred by the learner \textit{and} have low risk form a set of feasible hypotheses. Others in $\cH^{\mathrm{Ex}} \setminus \cH^{\mathrm{Fe}}$ are easily disfavored by the learner.}
  \label{fig:feasible_hypotheses}
\end{wrapfigure}

To begin with, we call the hypothesis space in the conventional sense \textit{expressible hypotheses}. 
$$\cH^{\mathrm{Ex}} \triangleq \{h\ |\ p(h) > 0\}$$

\vspace{-6pt}
Hypotheses associated with high likelihoods of data are referred to as \textit{low-risk hypotheses}, with a risk measure $\Rb$. 
$$\cH^\mathrm{Lr} \triangleq \{h \in \cH^{\mathrm{Ex}}\ |\ \EE_{d \sim \PP_{\mu}}[\Rb(h, d)] < \epsilon\}$$

Finally, viewing learning as search over a hypothesis space \cite{mitchell1997ML}, and viewing search as performing Bayesian inference \cite{Bayesian_for_NNs, dive_into_DL}, the learner would end up with a hypothesis with a high posterior probability, which is both \textit{a priori} preferred by the learner \textit{and} low-risk. Such hypotheses that are \textit{a posteriori} preferred form the set of \textit{feasible hypotheses}. 

\vspace{-6pt}
\begin{align*}
    \cH^\mathrm{Fe} \triangleq \{h \in \cH^{\mathrm{Ex}}\ |\ \EE_{d \sim \PP_{\mu}}[\PP(h\ |\ d)] > \gamma \} 
    = \{h \in \cH^{\mathrm{Ex}}\ |\ \EE_{d \sim \PP_{\mu}}[\frac{\PP(d\ |\ h)\PP(h)}{\PP(d)}] > \gamma \}
\end{align*}
\vspace{-6pt}

Note that being low-risk is a necessary condition for a hypothesis to be feasible, since a small $\Rb(h, d)$ is in line with a large $\PP(d|h)$. To reflect this correspondence, we can assume that the threshold $\gamma$ is always chosen such that $\cH^\mathrm{Fe} \subseteq \cH^\mathrm{Lr}$. Hereafter, we drop the superscript on feasible hypotheses unless noted otherwise, as feasible hypotheses are the most relevant in most contexts, i.e. $\cH \equiv \cH^\mathrm{Fe}$. In summary, for any $\cD_k$: $\cH_k \equiv \cH^\mathrm{Fe}_k \subseteq \cH^\mathrm{Lr}_k \subseteq \cH^\mathrm{Ex}_k$.

\textbf{Different domains $\cD_1, ..., \cD_k$ induce different $\cH_1, ..., \cH_k$}. When the learner is fixed, feasible hypotheses would depend on the data. Hence, including the subscripts for $\cH$ in accordance with the subscripts for $\cD$ reflects the possibility that feasible hypotheses are different between domains, regardless of whether they result from fundamentally distinct expressible hypothesis spaces. Similarly to the definition of $\cD_{\leq k}$, $\cH_{\leq k} \triangleq \cH_1 \times ... \times \cH_k$. When the focus is on the learning outcome rather than its dynamics, we can conceptually equate learning on $\cH^{\mathrm{Ex}}_k$ given $\cD_k$ with learning on $\cH_k$ because hypotheses in $\cH^{\mathrm{Ex}}_k \setminus \cH_k$ could be easily eliminated.

\paragraph{Expressivity, Learnability, and Generalizability}

Our notation builds consensus on formally stating expressivity, learnability and generalization, summarized in Tab~\ref{tab:ELG_Inv}. In multi-domain learning, all three notions depend on a central concept of invariance or invariance-capturing hypothesis, which can be conveniently expressed in terms of $\cH^\mathrm{Fe}$ and $\cH^\mathrm{Lr}$ introduced in $\S$~\ref{sec:Ex_Lr_Fe_Hypotheses}. Two important messages: (1) \textit{Expressivity does not imply learnability}. This can be precisely explained by the difference between $\cH^\mathrm{Fe}$ and $\cH^\mathrm{Lr}$: certain low-risk hypotheses might be unreachable by the optimization process or might be disfavored due to the learner's inductive bias. (2) \textit{Learnability and generalizability are interchangeable} because they share the same form: ``with high probability, expected risk is small" \cite{multisource_PAC}, where the probability is with respect to possible draws of a training set $d_k \sim \PP_{\mu_k}$.

\section{Difficulty Progression}
\label{sec:difficulty_progression}
\newcommand{\Succ}[0]{$\mathbf{Succ}$ }
\newcommand{\Cpx}[0]{$\mathbf{K}$}

Inductive generalization is achieved when the inferred rules or algorithms apply beyond the bounded set of observations from which they are learned. Current approaches to OODG typically partition the task space into only two parts: one in-domain and one out-of-domain. We advocate considering the task space as containing a series of domains. This has the advantages of (1) revealing the successorship among domains, (2) defining a temporal axis along which graceful degradation can be evaluated ($\S$~\ref{sec:DGR}), and (3) foreshadowing a capacity growth underlying optimal hypotheses, which can be exploited to induce inductive generalization ($\S$~\ref{sec:inductive_learnability}).

\paragraph{Conceptualizing the Successorship Among Domains}
\label{sec:peano}

\newcommand{\mybox}[4]{
    \begin{figure}[h]
        \centering
    \begin{tikzpicture}
        
        \node[anchor=text,text width=\columnwidth-1.2cm, draw=white, rounded corners, line width=.5pt, fill=#3, inner sep=5mm] (big) {\\#4};
        \node[draw=white, rounded corners, line width=.5pt, fill=#2, anchor=west, xshift=1mm] (small) at (big.north west) {#1};
    \end{tikzpicture}
    \end{figure}
}

Peano's axioms naturally support inductive definitions, which we leverage to define \textit{progressively difficult domains}. Consider a series of domains indexed by natural numbers, denoted by the fraktur letter $\mathfrak{D} = \{\cD_1, \cD_2, ..., \cD_k, ...\}$\footnote{
Without loss of generality, we index from 1 instead of 0 to maintain consistency of notation.
}. We say $\mathfrak{D}$ specifies an inductive problem if it, along with a data successor \Succ, satisfies Peano's axioms:

\vspace{-6pt}
\mybox{
    \small
    \textcolor{white}{$(\mathfrak{D}, \cD_1, \mathbf{Succ})$ specifies a model of the Peano axioms}
}{black!60}{gray!10}{
    \footnotesize
    \vspace{-6pt}
    \textcolor{MidnightBlue}{\textbf{1.} Unique origin}: $\mathcal{D}_1 \in \mathfrak{D}$
    
    \setlength{\parskip}{3pt}
    \textcolor{MidnightBlue}{\textbf{2.} $\mathfrak{D}$ is closed under \Succ}: If $\mathcal{D}_k\in \mathfrak{D}$, then $\mathcal{D}_{k+1} = \mathbf{Succ}(\mathcal{D}_k) \in \mathfrak{D}$

    \textcolor{MidnightBlue}{\textbf{3.} \Succ is bijective}: If $\mathcal{D}_k, \mathcal{D}_j \in \mathfrak{D}$, $\mathbf{Succ}(\mathcal{D}_k) = \mathbf{Succ}(\mathcal{D}_j)$ implies $\mathcal{D}_k = \mathcal{D}_j$.

    \textcolor{MidnightBlue}{\textbf{4.} No loop}: For every $\mathcal{D}$, $\mathbf{Succ}(\mathcal{D}) \neq \mathcal{D}_1$.

    \textcolor{MidnightBlue}{\textbf{5.} No junk / Axiom of Induction}: If $\mathfrak{A}$ is a set such that: $\cD_1 \in \mathfrak{A}$, every element in $\mathfrak{A}$ can be derived via applying \Succ a number of times to $\cD_1$, then $\mathfrak{A}$ contains every element in $\mathfrak{D}$.
    \vspace{-6pt}
}
\vspace{-6pt}

A few comments on how this formalism connects to practical cases are warranted. First, the “no junk” axiom critically implies that a testing sample cannot go out-of-domain in arbitrary ways. Any OOD instance should only differ from in-domain instances in a \textit{principled} way informed by \Succ. As such, one can only expect ``\textit{principled} inductively generalization", and cannot expect, for example, a model trained on mazes to generalize to poem-writing, unless non-trivial efforts have been dedicated to abstracting and unifying structure of both domains. We formalize such principles in $\S$~\ref{sec:nicenessS}. We note that formalizing ``task relatedness" is also an ongoing investigation in multi-task learning \cite{ben-david_multitask_relatedness, lifelongML}.

Second, $\mathfrak{D}$ is isomorphic to natural numbers\footnote{
Isomorphism is used in a much looser way in our context than in mathematics, because it is unclear how arithmetics or binary relations can be defined over domains. Our main aim is to draw analogies between how the inductive principle is embedded in the definition of natural numbers and how learning the inductive principle is vital for inductive generalization. 
},
which explains why in the literature “count” is such a pervasive concept involved in the definition of IND/OOD splits. Indeed, the most straightforward way to quantify complexity is to take advantage of a countable variable. Such countable variables could be tokens in a sequence \cite{NN_Chomsky_Hierarchy, lengen_arithmetic}, nodes in a graph \cite{clrs}, moves in search \cite{struggle_to_search, rubiks_cube_SSL}, depth of nested brackets \cite{circuit_complexity_hardattn, bounded_hierarchical_languages, learn_to_solve_recursively}, or empty entries in Sudoku \cite{causal_lm_logic_puzzles}. Note that the count variable does not have to correlate with input sizes.
For example, the depth of nesting structures or the number of moves in search can vary independently of input sizes, but they are apt to define $\mathfrak{D}$.

Third, generalization problems concerned with \textit{continuous} spaces fall out of scope. We pointed out challenges regarding a further unification in App~\ref{app:continuous_space} and delegate them to future studies.

\paragraph{Principled Difficulty Progression}

The structure required by Peano's axioms qualitatively characterizes the direction of generalization. However, it is mathematically unsolid because $\mathfrak{D}$ lacks a group structure with a binary operation. Therefore, this section quantitatively characterizes the difficulty gap between domains and the niceness of a successor function.  

\textbf{Difficulty of $\cD$} Following \citet{bengio2009curriculum}, we use entropy as a measure of difficulty. We require that the entropy of distributions ($\PP_{\mu_k}$) monotonically increases with $k$. Thus, \Succ must account for the amount of difficulty gain between successive domains, which is discussed next.

\textbf{Niceness Properties of \Succ} \label{sec:nicenessS}
Without formalizing niceness properties of $\mathbf{Succ}$, the definition of $\mathfrak{D}$ is inevitably vacuous because specifying an inductive problem would reduce to a game of intuitively finding orders among datasets. We need niceness restrictions on \Succ so that (1) \Succ directly reflects the difficulty gain between successive domains; (2) expectations to generalize in impossible ways\footnote{
It is impossible to transcend expressivity barriers. For instance, in language recognition, regular and context-free languages should never belong to the same $\mathfrak{D}$ without simplifying assumptions. And we should impose restrictions on \Succ to avoid that}
are clearly disallowed.

Since we generally assume that data are strings, \Succ can be realized as a list of probabilistic transducers $\{\Tb_1, \Tb_2, ...\}$. We say that $\Tb_k$ can generate $\cD_{k+1}$ from $\cD_k$ if it satisfies Eq.~\ref{eq:transducer}.

\vspace{-6pt}

\begin{equation}
    \forall b \in \cS_{k+1},\ \  
    \PP_{\mu_{k+1}}(b) = \frac{\sum_{a \in \cS_k} \PP_{\mu_k}(a)\PP[\Tb_k(a) = b]}{\sum_{c \in \cS_{k+1}}\sum_{a \in \cS_k} \PP_{\mu_k}(a)\PP[\Tb_k(a) = c]}
    \label{eq:transducer}
\end{equation}

The complexity of $\Tb_k$ quantifies the difficulty gap between $\cD_{k+1}$ and $\cD_k$. The complexity of a probabilistic transducer, \Cpx$(\Tb)$, can be measured by the totality of its alphabets, states, and transition rules. Then, niceness properties of \Succ can be defined via regulating the behavior of difficulty gaps. We impose two niceness properties. 
Informally, the first property requires a constant difficulty gap (in the limit) between consecutive domains; the second property requires that no subsequence of $\mathfrak{D}$ can have a difficulty gap (in the limit) lower than that of $\mathfrak{D}$. We formalize them in Definitions~\ref{def:const_gap} and \ref{def:no_simpler_subseq}.

\begin{definition}[Constant difficulty gap] \label{def:const_gap}
\textit{There exist $\Tb, \bar{k}$ such that $\Tb$ satisfies Eq.~\ref{eq:transducer} for all $k \geq\bar{k}$,
and \Cpx$(\Tb') \geq$\Cpx$(\Tb)$ for any other $\Tb'$ which also satisfies Eq.~\ref{eq:transducer} for some $k \geq\bar{k}$.}
\end{definition}

The second property is imposed contingent on that the first property holding, i.e. $\Tb, \bar{k}$ already exist.

\begin{definition}[No simpler subsequence] \label{def:no_simpler_subseq}
\textit{For all $\MM = \{i_1, i_2, ...\}$\footnote{
Having cardinality $\NN$ implies a bijection between $\MM$ and $\NN$. So elements of $\MM$ can be indexed by $\NN$.
} such that $\MM \subset \NN$ and $\MM$ has the same cardinality as $\NN$, 
$\nexists \Tb'$ which satisfies Eq.~\ref{eq:transducer} for all $k \in \{i\ |\ i \geq \bar{k}, i \in \MM\}$ and \Cpx$(\Tb')<$\Cpx$(\Tb)$.
}
\end{definition}

\section{Evaluation by Graceful Degradation}
\label{sec:DGR}

It is only worth discussing generalization when (multidomain) expressivity and learnability are no longer major issues. Therefore, we put forth the following assumptions before delving deeper.

\begin{assumption} [No issue with expressivity or learnability] \label{ass:easyExLr}
    $\forall k, |\bigcap_{j \leq k} \cH_j^{\mathrm{Lr}}| > 0$, and with high probability, 
    $L(d_k) \in \bigcap_{j \leq k} \cH_j \subseteq \bigcap_{j \leq k} \cH_j^{\mathrm{Lr}}$.\footnote{
        Future work can study the cases when $|\bigcap_{j \in cX} \cH_j^{\mathrm{Lr}}| > 0$ or $\bigcap_{j \in X} \cH_j \subseteq \bigcap_{j \in X} \cH_j^{\mathrm{Lr}}$ holds for certain subsets of $\NN$ ($\cX \in \NN$).
    }
\end{assumption}

\begin{assumption} [No issue with hard-to-easy generalization] \label{ass:easyH2E}
    If $L(d_k)$ is performant in $\cD_k$, then it is performant in lower-difficulty domains as well, i.e. $L(d_k) = \hat{h}^*_k \in \bigcap_{j=1}^k \cH_j$.
\end{assumption}

Assumption~\ref{ass:easyH2E} allows us to omit the distinction between $L(d_k)$ and $L(d_{\leq k})$ to avoid verbosity\footnote{
There are interesting questions should this assumption not hold \cite{hardtoeasy}, which follow-up studies can explore. 
}. Due to near perfect in-domain learnability, in-domain metrics cannot effectively distinguish different solutions trained to convergence, motivating a better metric focusing on the ability to generalize toward harder problems. To this end, we evaluate inductive generalization by \textit{degradation} (\textbf{DGR}), defined as a discounted sum of risks over harder domains $\mathcal{D}_{>k}$: 
\vspace{-12pt}

\begin{equation}
    \textbf{DGR}(h_k) = \sum_{m=k+1}^{\infty} \omega_m \mathbb{E}_{(x, y) \sim \mu_m}\Big[\Rb\big(h_k, (x, y)\big)\Big]
    \label{eq:DGR}
\end{equation}
\vspace{-9pt}

$\omega_m$'s are hyperparameters and $\sum_{m=k+1}^{\infty} \omega_m = 1$, allowing us to weigh near- and remote-future risks differently. A model exhibits \textit{graceful degradation} if its \textbf{DGR} is small.
 
\section{Inductive Learnability}
\label{sec:inductive_learnability}
\newcommand{\baselearner}[0]{$L^{\text{Base}}$ }
\newcommand{\indlearner}[0]{$L^{\text{Ind}}$ }
\newcommand{\lifelonglearner}[0]{$L^{\text{Life}}$ }
\newcommand{\prospectivelearner}[0]{$L^{\text{Pros}}$ }
\newcommand{\invlearner}[0]{$L^{\text{Inv}}$ }

\newcommand{\Ind}[0]{$\mathbf{Ind}$ }
\newcommand{\Indk}[0]{$\mathbf{Ind}_k$ }

We provide a formal definition of inductive learnability under the $(\epsilon, \delta)$-learning framework. We assume a base-level learner, \baselearner, which is able to perform PAC-learning within each individual $\mathcal{D}_i$. Then, we assume an inductive learner, \indlearner, which is a meta-level learner. We first define the functional forms of \baselearner and \indlearner, then define inductive-learnability based on the gain in graceful degradation of \indlearner over \baselearner. We are aware that ``induction" or ``inductive learning" have different interpretations, e.g., in classic machine learning \cite{mitchell1997ML, utgoff_ML_of_indbias} vs. cognitive psychology \cite{InductiveReasoning, stanford_Encyclopedia, tenenbaum1999}. To avoid confusion, inductive learning in this paper specifically refers to learning a successor function over models. We denote the model successor by \Ind to distinguish it from the data successor \Succ.

\paragraph{Base Learner}

\baselearner has functional form $\cF^{\text{Base}} = \{\cF_k^{\text{Base}}\ |\ k \in \NN\}$, where $\cF_k^{\text{Base}} \subseteq \{f_k: \cD_k \rightarrow \cH_k^{\text{Ex}}\}$ is the set of learning algorithms that accepts data in $\cD_k$ and yields a hypothesis in $\cH_k^{\text{Ex}}$. \footnote{
It is not a must that the base learner only access data from a single domain at a time. It is possible to have the base learner learn from data \textit{up to} $\cD_i$ at a time. However, we believe that this design choice matters less for presenting our framework at the high level. Thus, to avoid verbosity, we stick with the scenario where the base learner learns from a single domain at a time. 
} 

\paragraph{Inductive Learner}
\label{sec:indlearner}
When it comes to \indlearner, it is helpful to elaborate on how its input and output spaces are defined. Vital learning signals for \indlearner can be hosted in two progressions. One is the difficulty progression over domains, corresponding to an ordered set of datasets: $d_{\leq k} = (d_1, ..., d_k)$. The other is the capacity progression over optimal hypotheses, corresponding to an ordered set of hypotheses inferred by \baselearner: $\hat{h}^*_{\leq k} = (\hat{h}_1^*, ..., \hat{h}_k^*)$. Therefore, the input space of each $f_k \in \cF_k^{\text{Ind}}$ is one that contains all possible $d_{\leq k}$'s and $\hat{h}^*_{\leq k}$'s, that is, $\cD_{\leq k} \times \cH_{\leq k}$. 

The output of \indlearner should be \Indk, which operates over hypotheses such that given $h_i \in \cH_i$, \Indk$(h_i) \in \cH_{i+1}$. It is clear that \Indk belongs to a function space, that is, $\cH^{\cH}$.
Together, \indlearner has the functional form $\cF^{\text{Ind}} = \{\cF_k^{\text{Ind}}\ |\ k \in \NN\}$, where $\cF_k^{\text{Ind}} \subseteq \{f_k: \cD_{\leq k} \times \cH_{\leq k} \rightarrow \cH^{\cH}\}$. 

Note, the difficulty progression must be reflected in the model progression as a trend of capacity growth, which must be captured by \Indk. In this sense, the goal of \indlearner is to infer a model successor that embodies capacity growth.

\paragraph{Success Criterion for \indlearner}
\label{sec:DGR_and_indlearnability}

\textit{Degradation} for \Indk can be defined following Eq.~\ref{eq:DGR}:
\vspace{-15pt}

\begin{align*}
    \textbf{DGR}(\mathbf{Ind}_k, h_k) = \sum_{m=k+1}^{\infty} \delta_m \mathbb{E}_{(x, y) \sim \mu_m}\Big[\Rb\big(\tilde{h}_m, (x, y)\big)\Big],\quad 
    \tilde{h}_m = \underset{\text{apply m-k times}}{\text{Ind}_k\Big(\text{Ind}_k\big(...}(h_k)\big)\Big)
\end{align*}

The success of \indlearner is defined in terms of its \textbf{DGR} relative to \baselearner. This is common in PAC learning, where success is defined in terms of relative risk to a Bayes-optimal or random hypothesis. Moreover, this relative definition also avoids unnecessary complication of a problem when the base learner already performs well and renders \Ind useless (for further discussion, see $\S$~\ref{sec:not_evolvingOH}). 

\begin{definition}[Inductive learnability] 
    \textit{\indlearner $(\epsilon, \delta, k)$-inductively learns from $\cD_{\leq k}$ with respect to \baselearner whose sample complexity is $n$\footnote{
        More formally, we must also have $(\epsilon, \delta, n)$ for the learnability conditions of \baselearner, that is, given at least $n$ data samples, $\underset{d^n_k \sim \mu_k}{\PP}\Big[\Rb\big(\hat{h}^*_k, d^n_k\big)\Big] \leq \epsilon\Big] \geq 1-\delta$. For convenience of notation, we omit $\epsilon, \delta$ associated with base-learnability as they are identical to the PAC definition \cite{intro_computationalLT, Valiant_theory_of_learnable}
    }, if with probability $1-\delta$, $L^{\text{Ind}}(d^n_{\leq k}, \hat{h}_{\leq k}^*)$ outputs \Indk such that \Indk degrades $\epsilon$-more gracefully than $\hat{h}^*_k$, that is,}
\end{definition}
\vspace{-6pt}

$$\underset{d^n_1 \sim \mu_1, ..., d^n_k \sim \mu_k}{\PP}\Big[\textbf{DGR}(\hat{h}_k^*) - \textbf{DGR}(\mathbf{Ind}_k, \hat{h}_k^*) \geq \epsilon\Big] \geq 1-\delta$$

\textit{where $\hat{h}_i^* = $\baselearner$(d^n_i)$.} Without loss of generality, we assume $n$ upperbounds both the sample complexities for \baselearner learning on \textit{all} of the first $k$ domains ($L^{\text{Base}}(d^n_1)$, ..., $L^{\text{Base}}(d^n_k)$), and the sample complexity for $L^{\text{Ind}}(d^n_{\leq k}, \hat{h}_{\leq k}^*)$.

\begin{table*}[t]
\setlength{\belowrulesep}{0pt}
\setlength{\aboverulesep}{0pt}
\footnotesize
\centering
\begin{tblr}{
  width=1\textwidth, stretch=1.5,
  cell{2}{1-6} = {bg=brown9},
  cell{3}{1-6} = {bg=blue9},
  cell{4}{1-6} = {bg=teal9},
  cell{5}{1-6} = {bg=red9},
  colspec  = {X[l, 3.4cm]X[l, 3.7cm]|X[c, 0.6cm]X[c, 1.5cm]|X[c, 0.6cm]X[c, 1.6cm]}
  }
\toprule
\SetCell[r=1]{f,3.5cm}{\textbf{Learning Paradigm}} & \SetCell[r=1]{f,3.8cm}{\textbf{Subcases}} & \textbf{Evolve Model} & \makecell[ct]{\textbf{Capacity}\\ \textbf{Growth}} & \textbf{Evolve Data} & \makecell[ct]{\textbf{Complexity}\\ \textbf{Growth}}\\

\midrule
\SetCell[r=1]{m,3.5cm}{
a. Learning under distributional shift \S\ref{sec:not_evolvingOH}} & \makecell[l]{Transfer/Multitask learning \\ Domain adaptation \\ Domain generalization \S\ref{subsubsec:inv-capturing} \\ Zero-shot generalization \S\ref{subsubsec:ITS}} & No & No & Yes & \textcolor{gray}{Unnecessary} \\
\hline[white]

b. Lifelong learning \S\ref{subsubsec:lifelong} & \makecell[l]{Online/Streaming learning\\Continual learning} & Yes & Yes & Yes & \textcolor{gray}{Unnecessary} \\
\hline[white]

c. Prospective learning \S\ref{subsubsec:prospective} & \textcolor{gray}{Unexplored} & Yes & \textcolor{gray}{Unnecessary} & Yes & \textcolor{gray}{Unnecessary} \\
\hline[white]

d. Inductive learning (ours) & \textcolor{gray}{Unexplored} & Yes & Yes & Yes & Yes \\
\bottomrule
\end{tblr}
\caption{Taxonomy of learning paradigms  the evolvement in data and model.}
\label{tab:many_paradigms}
\end{table*}


\begin{table*}[t]
\def\arraystretch{1.5}
\centering
\footnotesize
\begin{tabular}{p{11pt}@{}p{0.35\textwidth}p{0.57\textwidth}@{}}
\toprule

\multicolumn{2}{l}{a. Learning under distributional shift \S\ref{sec:not_evolvingOH}} &  \SetCell[r=2]{h, 4cm}{\raisebox{-.5\height}{\includegraphics[width=5cm]{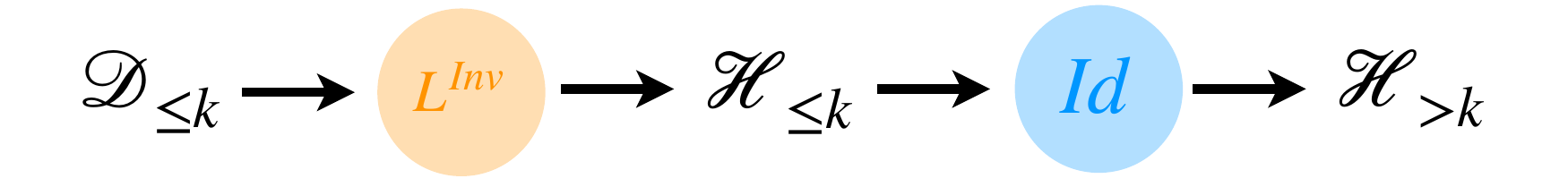}}} \\

\rowcolor{SpringGreen!60} 
\cmark 
& 1. $\exists \text{Inv}$, i.e. $| \bigcap_{j=1}^{\infty}\cH_j|>0$ 
 & 2. \text{Inv} shared by future distributions can be uniquely identified within a finite horizon. i.e. $\exists k$ s.t. w.h.p, $L^{\text{Inv}}(d_{\leq k}) \in \bigcap_{j=1}^{\infty} \cH_j$ \\
\rowcolor{Red!10}
\xmark
& 1. Universal $\text{Inv}$ does not exist: $\forall k, \epsilon>0$, $\exists m>k$ s.t. $|\bigcap_{j\leq k\ or\ j=m} \cH_j| < \epsilon$
& 2. Universal $\text{Inv}$ exists but disfavored by the learner lacking incentives: w.h.p $L^{Inv}(d_{\leq k}) \in \big(\bigcap_{j \leq k} \cH_j\big) \setminus \big(\bigcap_{m>k} \cH_m\big)$.
 \\
\midrule

\multicolumn{2}{l}{b. Lifelong learning \S\ref{subsubsec:lifelong}} & 
\raisebox{-.6\height}{\includegraphics[width=5cm]{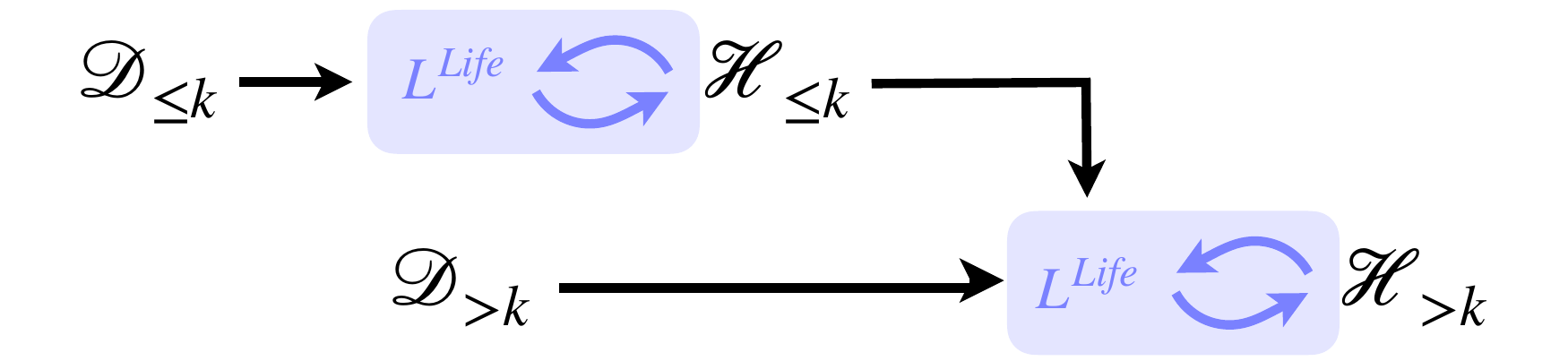}} \\

\rowcolor{SpringGreen!60} \cmark 
& \multicolumn{2}{p{0.93\textwidth}}{No predictable pattern that \textit{fully} account for data evolvement. Data of a new domain is always available.}\\
\rowcolor{Red!10}\xmark 
& \multicolumn{2}{p{0.93\textwidth}}{The volume of support expands combinatorially for unseen domains, in which \indlearner may help if the expansion is principled.} \\
\midrule

\multicolumn{2}{l}{c. Prospective learning \S\ref{subsubsec:prospective}}  &
\raisebox{-.5\height}{\includegraphics[width=5cm]{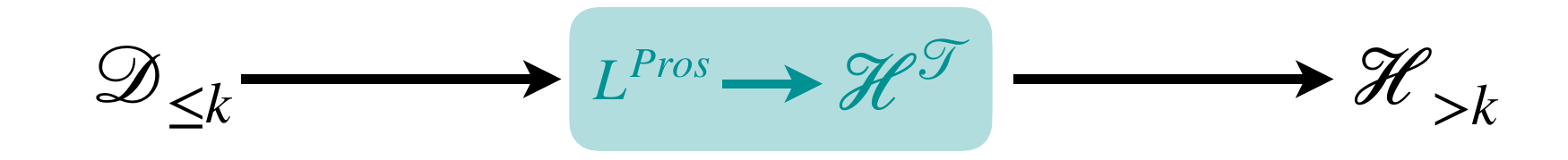}} \\
\rowcolor{SpringGreen!60} \cmark &  \multicolumn{2}{p{0.93\textwidth}}{Data is generated by a stochastic process indexed by time $t \in \cT$.} \\
\rowcolor{Red!10}\xmark & \multicolumn{2}{p{0.93\textwidth}}{The stochastic data-generating process cannot be identified within finite time. i.e. $\bar{t}$ required by Definition 2 in \citet{prospective} does not exist. In this case, \lifelonglearner may be more suitable.} \\
\midrule

\multicolumn{2}{l}{d. Inductive learning \S\ref{sec:inductive_learnability}} &
\SetCell[r=3]{m, 4.9cm}{\raisebox{-.5\height}{\includegraphics[width=5cm]{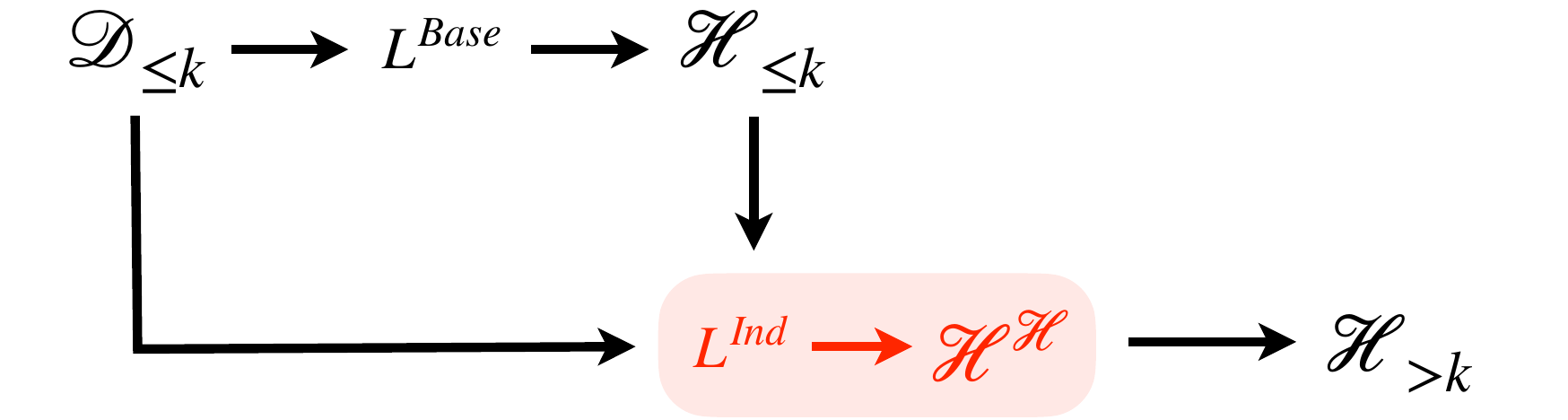}}} \\
\rowcolor{SpringGreen!60} \cmark & \multicolumn{2}{p{0.93\textwidth}}{Data is inductively generated by applying \Succ to some base case.} \\
\rowcolor{Red!10} \xmark & 
\multicolumn{2}{p{0.93\textwidth}}{
    1. \baselearner can already provably generalize: $\exists k$ s.t. w.h.p \baselearner$(d_{\leq k}) \in \bigcap_{j=1}^{\infty} \cH_j$, rendering \Ind needless. }\\
\rowcolor{Red!10}& \multicolumn{2}{p{0.93\textwidth}}{
    2. Difficulty gap does not converge to constant (violating Def~\ref{def:const_gap}). The data evolvement pattern can always go beyond what is possible to be captured during learning on $\cD_{\leq k}$.  In this case, use \lifelonglearner. 
}\\
\rowcolor{Red!10}& \multicolumn{2}{p{0.93\textwidth}}{
    3. $\mathfrak{D}$ has simpler subsequences (violating Def~\ref{def:no_simpler_subseq}).
    In this case, \prospectivelearner may help capture the transition between subsequences. 
}\\

\bottomrule
\end{tabular}
\caption{We clarify the differentiating factors between four learning paradigms with compact schematics. Each has advantage in certain scenarios that accord well with their core assumptions. We use shorthands ``w.h.p" for ``with high probability" and ``$\text{Inv}$" for ``invariance". Suitable conditions are marked \cmark, while unsuitable lines are indicated with \xmark.}
\label{tab:schematic}
\end{table*}


\section{Relation to Existing Learning Frameworks}
\label{sec:learning_paradigms}

\paragraph{The Need for An Evolving Optimal Hypothesis}
\label{sec:need_evolving_h}

Learning paradigms differ in the interplay between receiving new data and inferring new hypotheses. We provide an overview with schematics in Tab~\ref{tab:schematic} and elaborate on how these compact schematics are derived in App~\ref{app:diagrams}. In this regard, a larger holistic paradigm, in which an optimal hypothesis is inferred once and does not evolve, encompasses numerous sub-frameworks. We name it \textit{learning under distributional shift}, with the shorthand \invlearner for the corresponding learner (Tab~\ref{tab:many_paradigms}a). Inductive learning reduces to this case when \Ind is the identity function ($\mathbf{Id}$). Generalization to new domains relies on the assumption that the invariances \cite{closer_look_graph_OODG} of training and unseen domains have non-trivial intersections. (Tab~\ref{tab:schematic}a). The methods by which the current \invlearner literature tackles OODG fall into two broad categories: generalization by capturing invariance and generalization by inference-time scaling. App~\ref{sec:not_evolvingOH} surveys both categories and explains how inductive learning should progress in light of their achievements and obstacles.

The hope for the OODG ability of an \invlearner (Tab~\ref{tab:schematic}a)\footnote{
Note, comprehensive deep learning theories for the statement ``w.h.p $L^{Inv}(d_{\leq k}) \in \big(\bigcap_{j \leq k} \cH_j\big) \setminus \big(\bigcap_{m>k} \cH_m\big)$" remain elusive, despite a few attempts \cite{GOTU, SGD_extreme_SB} and abundant empirical evidence \cite{fate_faith, shortcuts}. Establishing impossibility theorems \cite{impossibility_domainadapt} by quantifying how simplicity biases constrain $\cH^{\mathrm{Fe}}$ relative to $\cH^{\mathrm{Lr}}$, thus causing $L^{\mathrm{Inv}}$'s failure on OODG, is a essential path forward. 
}
can break when either there is no invariance or the invariance is disfavored by the learner (e.g. via a simplicity bias\footnote{
In theoretical AI, ``Occam's razor" \cite{mackay2003info_infr_learning, Hutter_universalAI, universal_solomonoff} refers to a universal simplicity bias.}
). 
Simplicity can be imposed by architecture
\cite{low_sensitivity_sparseboolean, low_sensitivity, low_LempelZiv_complexity}, optimization algorithms \cite{GD_min_norm, optimizer_low_nuclearnorm_divergence_maxmargin, SGD_extreme_SB}, or both \cite{spectral_bias, low_frequency}. 

To overcome the limit of a static optimal hypothesis, the optimal hypothesis must evolve along with the distributional shift. This is captured by the general case of our framework, where \Ind$\neq \mathbf{Id}$. Lifelong learning (LL) \cite{lifelongML}, prospective learning (PL) \cite{prospective} and inductive learning (IL) (Tab~\ref{tab:many_paradigms} bcd) share the characteristic of an evolving optimal hypothesis, lending themselves to a future-oriented objective.\footnote{
    Although \citet{prospective} characterized LL as retrospective as opposed to prospective, \citet{LLconstrainedRL} argued that LL can be regarded as optimizing an infinite-horizon reward subject to informational constraints.
} In fact, LL, PL and IL are equivalent up to syntactic transformations over their graphical representations (App~\ref{app:diagrams}). However, we are not suggesting a replacement. LL, PL, and IL put different emphasizes on the form of predictability underlying data evolvement (Tab~\ref{tab:schematic} bcd), which will crucially shape modeling considerations. Uniquely in IL is the difficulty progression, with formal assumptions about how consecutive difficulty levels are related (\S~\ref{sec:nicenessS}). We believe that LL, PL and IL have distinct strengths, which we discuss next to aid practitioners in their decision-making.

To better motivate this section, we note that many empirical studies on zero-shot generalization \cite{pitfall_NTP, cogpsy_gpt3, fate_faith, symbolic_metaprogram_search, struggle_to_search, symbolic_brittleness, spatial_understanding, learn_to_solve_recursively, zhou2024lengen} are implicitly situating themselves in the learning paradigm for \invlearner, where it must hold that the model has been pretrained on $\cD_{\leq k}$ for a sufficiently large $k$ so that the intersection of future low-risk hypotheses has been identified. The implicit commitment to such assumptions without justification has led to a proliferation of negative results where the attribution of failure is ambiguous. We argue that many of these negative results are a reflection more of the mismatch between characteristics of the problem and the learning paradigm chosen, than of the fundamental incompetence in individual realizations of \invlearner. We intend to call for a rigorous examination of assumptions tied to the model (hypothesis spaces) and the model's past training data in future OODG research \cite{embers}. To facilitate this effort, we differentiate the comparative strengths of various learning paradigms with consistent terminology (Tab~\ref{tab:schematic}).

\paragraph{Lifelong Learning}
\label{subsubsec:lifelong}
\citet{manyPAC} standardized and hierarchically organized many learning problems under the PAC framework.  We inherit and extend their taxonomy with an organizational overview in Tab~\ref{tab:many_paradigms} and detailed graphical illustrations in App~\ref{app:diagrams}. According to \citet{manyPAC}, a lifelong learner, \lifelonglearner, has the functional form $\cF^{\text{Life}} = \{\cF_k^{\text{Life}} | k \in \NN\}$, where $\cF_k^{\text{Life}} \subseteq \{f_k: \cD_k \times \cH_{k-1} \mapsto \cH_k\}$.

Comparing \lifelonglearner and \indlearner, the crucial benefit of \Indk is that it eschews the need for data from a higher difficulty level, whereas \lifelonglearner only works if new data are available. However, we do not mean to render LL inferior to IL. The fundamental characterizing aspect of LL is the assumption that \textit{no} predictable patterns can \textit{fully} account for data evolvement, necessitating \textit{perpetual adaptation}. Any attempt to remove the dependency $\cD_{> k} \longrightarrow$ \lifelonglearner is essentially a departure from LL to other learning paradigms. 
On the other hand, if attempts fail to well define the difficulty progression of IL or the stochastic process of PL, there could be a chance that the problem can be handled by LL (Tab~\ref{tab:schematic} b).

\definecolor{lightgray}{RGB}{230,230,230}
\newcommand{\python}[1]{\colorbox{lightgray}{\color{red}{\texttt{#1}}}}

\begin{table*}[h]
\setlength{\belowrulesep}{0pt}
\setlength{\aboverulesep}{0pt}
\scriptsize
\begin{tblr}{
  width=1\textwidth, stretch=1.5,
  cell{1-11}{1} = {bg=brown6, fg=white},
  cell{1-11}{2} = {bg=brown8},
  cell{1-11}{3} = {bg=brown9},
  cell{1-11}{4-5} = {bg=red9},
  colspec  = {X[l, 1.45cm]X[l, 1.85cm]X[l, 2cm]X[l, 3.8cm]X[l, 2.75cm]}
  }
\toprule
\textbf{Roadmap} & \textbf{Our formulation} & \textbf{Pressing questions} & \textbf{Historical insights} & \textbf{Required adaptations} \\
\midrule
\textbf{1. Task} & Learn \Ind & Provable guarantees & Theories assuming support mismatch & Quantify divergence of $\hat{h}^*_k$ \\
\hline[white]

\textbf{2. Experience} & Training signals lie in $\cD_{\leq k} \times \cH_{\leq k}$ & Extract/enrich training signals & \textbf{BMA}: Multiple compelling ``moments" of $\hat{h}^*_k$ & Operationalize the curation of training signals \\
\hline[white]
\SetCell[r=5]{m, 2.1cm}{\textbf{3. Represent} \\ \textbf{Target} \\ \textbf{Functions}} & \SetCell[r=5]{m, 2.7cm}{None} & \SetCell[r=5]{m, 3.3cm}{Representations of \\$\cH$ ($h$) and $\cH^{\cH}$($\mathbf{Ind}$)} & \textbf{MPL}: Metaprograms revise programs & Connectionist counterpart \\
    \cline[white]{4-5}
    & & & \textbf{NAS}: Encode the syntax of $h$ & Encode mutation of syntaxes\\
    \cline[white]{4-5}
    & & & \textbf{Differentiable NAS}: \Ind is vector arithmetic & Learn the optimal \Ind \\
    \cline[white]{4-5}
    & & & \textbf{EA+NAS}: $f(\hat{h}^*_k) = \hat{h}^{\text{Init}}_{k+1}$ & Directly output $\hat{h}^*_{k+1}$ \\
    \cline[white]{4-5}
    & & & \textbf{CL}: Subspaces of $\cH^{\cH}$ that induce capacity growth & Align  data progression and capacity growth \\
    \cline[white]{4-5}
    & & & \textbf{Adapters}: Low-rank approx. of $\cH^{\cH}$ & Adapters that embody \Ind \\
    
\hline[white]  
\textbf{4. Metric} & Graceful degradation & Surrogates for practical use & None & None \\

\hline[white]
\textbf{5. Learning Mechanism} & None & Gradient descent vs. other algorithms & \textbf{MPL}: Bayesian inference & Hybrid it into a neurosymbolic system \\

\bottomrule
\end{tblr}
\caption{Existing techniques developed to address seemingly irrelevant questions can be repurposed to learn model successors in practice. \textbf{BMA}: Bayesian Model Averaging. \textbf{MPL}: Metaprogram Learner. \textbf{NAS}: Neural Architecture Search. \textbf{EA}: Evolutionary Algorithms. \textbf{CL}: Curriculum Learning}
\label{tab:design_indlearner}
\end{table*}

\paragraph{Prospective Learning}
\label{subsubsec:prospective}

\citet{prospective} argues that most learning problems can be characterized as \textit{retrospective} learning, because they focus on \textit{adapting} to new tasks rather than actively \textit{anticipating} task shifts. Hence, \citet{prospective} defines \textit{prospective} learning as a complement to \textit{retrospective} learning, where the learner takes as input a sequence of time-indexed datasets and outputs a sequence of time-indexed hypotheses. According to \citet{prospective}, a prospective learner, \prospectivelearner, has the functional form $\cF^{\text{Pros}} = \{\cF_k^{\text{Pros}} | k \in \NN\}$, where $\cF_k^{\text{Pros}} \subseteq \{f_k: \cD^{\cT} \mapsto \cH^{\cT}\}, \cT = \{1, 2, ..., t, ...\}$. Note, $\cD^{\cT}$ denotes a function space, which is the set of functions that map from time indices to datasets. Similarly, each element in the function space $\cH^{\cT}$ is a time-indexed sequence of hypotheses. PL assumes that the time-indexed data are generated by an (unknown) stochastic process. 

PL and IL both argue that predictable patterns cannot be captured (or even revealed) if one sticks to a fixed $\cH^{\mathrm{Ex}}$ (as in \invlearner), or only allows for additive expansion of $\cH^{\mathrm{Ex}}$ (as in \lifelonglearner; Fig~\ref{fig:cont_life}). Instead, the search for a solution should take place in a higher-order space which is combinatorially larger than the primitive $\cH^{\mathrm{Ex}}$. In PL, such a higher-order space is $\cH^{\cT}$, and in IL, it is $\cH^{\cH}$.

\section{Historical Insights for Defining \indlearner}
\label{sec:roadmap}

\citet{mitchell1997ML} states that building a learning system requires specifying a \textit{task}, an \textit{experience}, and a \textit{performance metric} at the design level, and then specifying a \textit{target function representation} and a \textit{learning mechanism} at the implementation level. These steps are outlined in Tab~\ref{tab:design_indlearner}, with the target function representation split into two sub-steps. The two right columns summarize techniques that can be borrowed from existing literature, together with proposed adaptation directions. A much more involved discussion is continued in App~\ref{app:roadmap_cont}. The character of our arguments is inspirational rather than instructive. The message we hope to convey is that, though the research territory we formalized here is underexplored, we do not have to chart a new landscape from scratch. Insights originated from nearby fields, which initially addressed seemingly disparate questions, can shed light on our goals. We hope that this paper will have profound implications on how a multidisciplinary endeavor can rejuvenate ``entrenched" wisdoms, and promote a shared understanding of the vast area they span. 

\section{Discussion}


Our point of view elucidates issues that may have received less focus in earlier studies, such as a) distinguishing feasible/expressible/low-risk hypotheses and b) the importance of justifying assumptions behind the choice of a learning paradigm. Several fundamental themes have surfaced, including evolving hypotheses, two levels of inference, and the synergy between data and model progressions, all pointing to the need for model successor functions. 
This work does not amount to a full-fledged theory of inductive generalization, but points to the kind of information we need to fill in.
Currently missing from our formalization is the principle by which the best timing to terminate \Ind can be decided. This question hinges on uncertainty quantification and the prediction of domain boundaries, where Bayesian deep learning \cite{position_BDL, the_case_for_BDL} may unlock future possibilities. We conclude with the final message that our field will benefit from integrating interdisciplinary insights to achieve the deep learning counterpart of ``inductive leap".

\newpage

\bibliography{main}
\bibliographystyle{neurips_2025}


\appendix

\newpage
\appendix
\onecolumn

\setcounter{figure}{0}
\renewcommand{\thefigure}{A\arabic{figure}}
\renewcommand{\theHfigure}{A\arabic{figure}}
\setcounter{table}{0}
\renewcommand{\thetable}{A\arabic{table}}
\renewcommand{\theHtable}{A\arabic{table}}

\section*{Learning Model Successors: Supplementary Material}

\section{Empirical Studies to Motivate and Exemplify}
\label{app:motivating}

Using the task of recognizing $dyck_1$ as a case study, this section empirically justifies the need for Inductive Learning by (1) illustrating a difficulty progression that demands inductive generalization, (2) exposing the limit of existing learning paradigms, and (3) presenting a realization of the base-learner and the inductive-learner. We demonstrate how the resulting model successor, $\mathbf{Ind}$, substantially improves generalization along the difficulty progression. 

\paragraph{Setup} $dyck_1$ is a well-known context-free language (balanced brackets), which can be generated recursively from a generative grammar. Here, the subscript $_1$ indicates that there is a single type of bracket ``()", which is the simplest case of $dyck$. Valid sequences can be generated via a generative grammar with a single nonterminal,
$S$, and three production rules ($\epsilon$ means the empty string). Since only the second production rule increases the nesting depth, we can control the maximum nesting depth in a training set by controlling how many times the second rule is called.

\begin{enumerate}[leftmargin=1cm, itemsep=0mm]
    \item $S \rightarrow \epsilon$\quad\quad 2. $S \rightarrow (\ S\ )$\quad\quad 3. $S \rightarrow S\ S$
\end{enumerate}

Invalid sequences are generated by corrupting valid sequences via one of the following steps.

\begin{enumerate}[leftmargin=1cm, itemsep=0mm]
    \item randomly delete a `(' or a `)'
    \item randomly insert a `(' or a `)'
    \item randomly substitute a `(' with a `)' or the other way around
    \item pick two valid sequences $X, Y$, concatenate them ``$X\ )\ Y$", then randomly insert `(' into $Y$
\end{enumerate}

Let $dyck_{1-m}$ denote $dyck_1$ language with nesting depth bounded by $m$. Hence, training data for $dyck_{1-m}$ will only include valid sequences with depth $\leq m$. The data for each $m$ is constructed with 50\% valid and 50\% invalid sequences and randomly split into 90\% train vs. 10\% val. 
We train RNNs\footnote{
We choose RNNs over Transformers for this experiment because Transformer has not yet overcome the expressivity issue for tasks that require sequential processing over the input. The number of Transformer layers has to grow logarithmically with the input length \cite{transformer_formal_language, positional_attn, hahn_limits_of_SA, shortcuts, transformers_logarithmic_depth}. Therefore, the Transformer is a candidate less capable than the RNN for modeling $dyck_1$.
} 
to classify valid vs. invalid sequences. All RNNs have one layer with 16 hidden units\footnote{
One may ask whether the reported generalization failure to greater depth can be simply due to insufficient model size or hyperparameter tuning. We posit that it is unlikely since multiple groups of researchers have reached a similar conclusion on the difficulty of generalizing to a greater depth, both for RNNs \cite{rnn_recognize_hierarchical_lang, rnn_nested_subj_verb} and self-attention models\cite{dyck_attn_seq2seq, learn_to_solve_recursively}
}. Batch size = 32, training steps = 15k. All results are averaged over 5 seeds.

\paragraph{Principled Difficulty Progression} The difficulty progression is induced by the nesting depth. We verify that the input distributions of our empirically generated datasets exhibit an increase in entropy (Tab~\ref{tab:dyck_entropy}). This is anticipated because as the size of support grows along the difficulty progression, entropy will also increase unless the distribution is highly skewed. Following \S\ref{sec:difficulty_progression}, we show that the data successor is well-behaving because the difficulty gap can be fully characterized by the probabilistic finite state transducer (PFST) shown in Fig~\ref{fig:PFST}.

\vspace{6pt}

\begin{adjustbox}{valign=c}
\begin{minipage}[t]{.55\linewidth}
    \centering
    \scriptsize
    \vspace{3pt}
        \begin{tabular}{@{}lcccc@{}}
        Entropy & $dyck_{1-1}$ & $dyck_{1-2}$ & $dyck_{1-3}$ & $dyck_{1-4}$ \\
        \midrule
        Non-cumulative & 6.50 & 9.41 & 9.96 & 10.20 \\
        Cumulative & 6.50 & 8.61 & 9.63 & 10.39 \\
        \midrule
        \end{tabular}
    \captionof{table}{\footnotesize Entropy of input distributions based on empirical frequency. Non-cumulative: estimated with val data at exactly level $m$. Cumulative: estimated with combined val data for levels $\leq m$. Entropy monotonically increases in both cases.}
    \label{tab:dyck_entropy}
\end{minipage}
\end{adjustbox}
\hfill
\begin{adjustbox}{valign=c}
\begin{minipage}[t]{.43\linewidth}
    \centering
      \scriptsize
      \begin{tikzpicture}
        [->, node distance = 1cm]
        \node[state, initial, accepting] (q0) {$q_0$};
        \node[state, right of=q0, xshift=2.3cm] (q1) {$q_1$};
        \draw 
        (q0) edge[loop above] node{$\star$:$\star$ ($p=x$)} (q0)
        (q0) edge[bend left, above] node{$\star$:$\star$ ($p=1-x$)} (q1)
        (q1) edge[bend left, below] node{$\epsilon$:``()" ($p=1$)} (q0);
      \end{tikzpicture}
  \captionof{figure}{\footnotesize A constant-size PFST that translates any sequence in the $dyck_m$ dataset into a sequence in the $dyck_{m+1}$ dataset. ``$\star:\star$" denotes the transition rule that consumes any symbol in its alphabet $\Sigma=$ \{`(', `)'\} and outputs the same symbol.}
  \label{fig:PFST}
\end{minipage}
\end{adjustbox}

\paragraph{Experiment 1} We first demonstrate that expressivity, learnability and generalizability are distinct problems. This motivates the need to distinguish expressible from feasible hypotheses \S~\ref{sec:Ex_Lr_Fe_Hypotheses}. Tab~\ref{tab:rnn_dyck} shows that RNNs have no difficulty learning to recognize $dyck_{1-m}$ for $m=1,2,3,4$, but cannot generalize to greater depth if the corresponding valid sequences were not seen during training. 

\vspace{6pt}

\begin{adjustbox}{valign=c}
\begin{minipage}[t]{.49\linewidth}
    \centering
    \scriptsize
        \begin{tabular}{@{}lcccc@{}}
         & & & & \\
        \textbf{Testing} $\Rightarrow$ & $dyck_{1-1}$ & $dyck_{1-2}$ & $dyck_{1-3}$ & $dyck_{1-4}$ \\
        \toprule
        
        \textbf{Training} $\Downarrow$ & & & & \\
        $dyck_{1-1}$ & 100 & 57 & 54 & 50 \\
        $dyck_{1-2}$ & 100 & 100 & 55 & 50 \\
        $dyck_{1-3}$ & 100 & 100 & 100 & 66 \\
        $dyck_{1-4}$ & 100 & 100 & 100 & 100 \\
        
        \bottomrule
        \end{tabular}
    \captionof{table}{\footnotesize RNN cannot generalize to greater depth on recognizing $dyck_{1-m}$. This is neither an expressivity nor a learnability issue because when sequences with greater depth are included for training, RNNs with the same capacity can fit well.}
    \label{tab:rnn_dyck}
\end{minipage}
\end{adjustbox}
\hfill
\begin{adjustbox}{valign=c}
\begin{minipage}[t]{.49\linewidth}
    \centering
    \scriptsize
        \begin{tabular}{@{}lccc@{}}
        \textbf{Testing} $\Rightarrow$ & \makecell{Same length\\Same depth} & \makecell{Double length\\Same depth} & \makecell{Triple length\\Same depth} \\
        \toprule
        
        \textbf{Training} $\Downarrow$ & & & \\
        $dyck_{1-1}$ & 100 & 100 & 100 \\
        $dyck_{1-2}$ & 100 & 100 & 100 \\
        $dyck_{1-3}$ & 100 & 100 & 100 \\
        $dyck_{1-4}$ & 100 & 100 & 100  \\
        
        \bottomrule
        \end{tabular}
    \captionof{table}{\footnotesize The nesting depth, rather than the input length, is the true difficulty indicator for $dyck_1$, because RNNs can length-generalize when controlling the depth. Training inputs have lengths up to 20. Double length = 40, triple length = 60.}
    \label{tab:rnn_lengthy}
\end{minipage}
\end{adjustbox}

\vspace{6pt}

The real cause of the generalization failure is the lack of incentive to settle for a more complex hypothesis without seeing more difficult instances. Recognizing $dyck_{1-m}$ requires the simulation of a counter that tracks the nesting depth. A model trained on $dyck_{1-m}$ will only develop $m$ counter states and has no incentive to develop more, even though the hypothesis space is theoretically able to express more \cite{hahn_limits_of_SA, transformers_logarithmic_depth}. The inability to develop counter states more than necessary is the true barrier while the input length is an artificial barrier, since Tab~\ref{tab:rnn_lengthy} shows that RNNs can generalize to much longer sequences as long as the nesting depth remains inside the training range.

\paragraph{Experiment 2} Our second experiment demonstrates that a continual/lifelong learning setting --- organizing training into distinct, easy-to-hard episodes --- does not enable generalization to greater depth (Tab~\ref{tab:rnn_continual}). Although lifelong learning allows one to evolve the optimal hypotheses, there is no transition between hypothesis classes. Thus, the argument still holds that there is a lack of incentive for converging at a hypothesis more complex (i.e. simulating more than $m$ counter states) than what is necessary to fit the training set.

\begin{table}[h]
\centering
\scriptsize
\begin{tabular}{@{}lccccc@{}}
\textbf{Testing} $\Rightarrow$ & $dyck_{1-1}$ & $dyck_{1-2}$ & $dyck_{1-3}$ & $dyck_{1-4}$ & $dyck_{1-5}$ \\
\toprule

\textbf{Training} $\Downarrow$ & & & & \\
$dyck_{1-1}$ & 100 & 100 & 54.8 & 50.2 & 49.8 \\
$dyck_{1-1,2,3}$ & 99.8 & 94.6 & 98.0 & 50.8 & 49.0 \\
$dyck_{1-1,2,3,4}$ & 100 & 100 & 100 & 99.8 & 60.0 \\

\bottomrule
\end{tabular}
\vspace{6pt}
\caption{$dyck_{1-a,b,c,d}$ means training follows the order: 10k steps on $dyck_{1-a}$, 10k steps on $dyck_{1-b}$, 10k steps on $dyck_{1-c}$, and 10k steps on $dyck_{1-d}$.}
\label{tab:rnn_continual}
\end{table}

\paragraph{Experiment 3} We leverage the $dyck_1$ task to showcase a successful realization of model successors. The key idea of learning model successors is learning at two levels of abstraction, necessitating a transition between hypothesis classes. To remind the reader, \baselearner captures regularities in data at/below each static difficulty level $(d_1, ..., d_k)$, yielding $(\hat{h}^*_1, ..., \hat{h}^*_k)$. Then, \indlearner captures regularities in models, yielding \Indk that can produce $\tilde{h}^*_m$ for $m>k$ without seeing any $d_m$. 

Following the previous two experiments, let $d_k$ correspond to the training set for $dyck_{1-k}$ and let $\hat{h}^*_k$ be the RNN that perfectly fits $dyck_{1-k}$. We will need to re-represent those RNNs into a proper input format for \indlearner. Leveraging the established theory that RNNs and finite state automata (FSAs) have computational correspondence, the literature has developed techniques to extract finite automata from RNN weights\footnote{https://github.com/DES-Lab/Extracting-FSM-From-RNNs} \cite{extract_FSA, FSA_decoded_from_RNNs, rule_extraction_from_rnn}. Tab~\ref{tab:FSAs} shows the extracted FSAs and their symbolic encodings. To encode each FSA, begin at the initial state and append all transition rules in order, separating them with the symbol `\#'. For brevity, transitions that lead to rejection are omitted (e.g. consuming `(' at $q_0$ will lead to rejection since there is no corresponding transition rule).

The task of learning model successors --- inferring $\hat{h}^*_{k+1}$ from $\hat{h}^*_{k}$ --- can be naturally formulated as language modeling. We randomly choose letters from [a-zA-Z] to name the states to avoid enforcing a particular order among the state names. We note that $q_0$ is always the accepting state, and that the transition rules of level $k$ are always contained in the transition rules of level $k+1$. Hence, each training datum for \indlearner can be constructed by concatenating the representation of the current hypothesis $\hat{h}^*_{k}$ with the additional transition rules for building $\hat{h}^*_{k+1}$, separating them with `<sep>', and terminating the sequence with `<eos>'. We use `<ns>' to denote a new state. Therefore, the vocabulary is [a-zA-Z, (, ), '\#',<ns>, <sep>, <eos>]. Tab~\ref{tab:FSAs} shows example training sequences. We train RNNs\footnote{One layer, hidden = 64, dropout = 0.1, batch = 32, training steps = 300, lr = 0.01, wd = 0.01.
} to become model successors, which in this setting are essentially decoder-only language models. Cross-entopy loss is applied to tokens succeeding `<sep>'. 

Our result indicates that training on merely three inductive steps ($\hat{h}^*_{1} \rightarrow \hat{h}^*_{2}$, $\hat{h}^*_{2} \rightarrow \hat{h}^*_{3}$, $\hat{h}^*_{3} \rightarrow \hat{h}^*_{4}$) enables perfect generalization up to $\hat{h}^*_{51} \rightarrow \hat{h}^*_{52}$. In terms of our success criteria defined in \S\ref{sec:DGR_and_indlearnability}, we achieve $\textbf{DGR}(\mathbf{Ind}_3, h_3) = 0$, in which $\delta_m=1$ if $4\leq m \leq 52$ and $=0$ otherwise\footnote{
Since the vocabulary allows for at most 52 distinct state names, we cannot test beyond $\hat{h}^*_{52}$
}. 
We obtain similar results even when the transition rules of $\hat{h}^*_{k}$ in each training sequence are shuffled. The model correctly learns that it is supposed to spot the state that has not been followed by a `(' from the prefix preceding `<sep>', and use it when generating the continuation. 

\begin{table}[h]
\centering
\scriptsize
\begin{tabular}{@{}cll@{}}
& \textbf{FSA extracted from RNN weights} & \makecell[lb]{\textbf{Symbolic encoding}\\ \textcolor{olive}{\textbf{Example training instances for \indlearner}}} \\
\toprule

$\hat{h}^*_1$ &
\parbox[c]{5cm}{\begin{tikzpicture}
[->, node distance = 0.35cm]
\node[state, initial, accepting] (q0) {$q_0$};
\node[state, right of=q0, xshift=0.5cm] (q1) {$q_1$};
\draw 
(q0) edge[bend left, above] node{(} (q1)
(q1) edge[bend left, below] node{)} (q0);
\end{tikzpicture}}
& \makecell[lc]{0 \# 0 ( 1 \# 1 ) 0 \\ \textcolor{olive}{N/A}} \\
\midrule

$\hat{h}^*_2$ &
\parbox[c]{5cm}{\begin{tikzpicture}
[->, node distance = 0.35cm]
\node[state, initial, accepting] (q0) {$q_0$};
\node[state, right of=q0, xshift=0.5cm] (q1) {$q_1$};
\node[state, right of=q1, xshift=0.5cm] (q2) {$q_2$};

\draw 
(q0) edge[bend left, above] node{(} (q1)
(q1) edge[bend left, below] node{)} (q0)
(q1) edge[bend left, above] node{(} (q2)
(q2) edge[bend left, below] node{)} (q1);
\end{tikzpicture}}
& \makecell[lc]{0 \# 0 ( 1 \# 1 ) 0 \# 1 ( 2 \# 2 ) 1 \\ \textcolor{olive}{a \# a ( c \# c ) a <sep> \# c ( <ns> \# <ns> ) c <eos>} \\
\textcolor{olive}{z \# z ( d \# d ) z <sep> \# d ( <ns> \# <ns> ) d <eos>} \\
\textcolor{olive}{i \# i ( p \# p ) i <sep> \# p ( <ns> \# <ns> ) p <eos>} }
\\
\midrule

$\hat{h}^*_3$ &
\parbox[c]{5cm}{\begin{tikzpicture}
[->, node distance = 0.35cm]
\node[state, initial, accepting] (q0) {$q_0$};
\node[state, right of=q0, xshift=0.5cm] (q1) {$q_1$};
\node[state, right of=q1, xshift=0.5cm] (q2) {$q_2$};
\node[state, right of=q2, xshift=0.5cm] (q3) {$q_3$};
\draw 
(q0) edge[bend left, above] node{(} (q1)
(q1) edge[bend left, below] node{)} (q0)
(q1) edge[bend left, above] node{(} (q2)
(q2) edge[bend left, below] node{)} (q1)
(q2) edge[bend left, above] node{(} (q3)
(q3) edge[bend left, below] node{)} (q2);
\end{tikzpicture}}
& \makecell[lc]{0 \# 0 ( 1 \# 1 ) 0 \# 1 ( 2 \# 2 ) 1 \# 2 ( 3 \# 3 ) 2 \\
\textcolor{olive}{b \# b ( s \# s ) b \# s ( k \# k ) s <sep> \# k ( <ns> \# <ns> ) k <eos>} \\
\textcolor{olive}{l \# l ( m \# m ) l \# m ( s \# s ) m <sep> \# s ( <ns> \# <ns> ) s <eos>} \\
\textcolor{olive}{o \# o ( d \# d ) o \# d ( g \# g ) d <sep> \# g ( <ns> \# <ns> ) g <eos>} \\
}\\
\midrule

$\hat{h}^*_4$ &
\parbox[c]{5cm}{\begin{tikzpicture}
[->, node distance = 0.35cm]
\node[state, initial, accepting] (q0) {$q_0$};
\node[state, right of=q0, xshift=0.5cm] (q1) {$q_1$};
\node[state, right of=q1, xshift=0.5cm] (q2) {$q_2$};
\node[state, right of=q2, xshift=0.5cm] (q3) {$q_3$};
\node[state, right of=q3, xshift=0.5cm] (q4) {$q_4$};
\draw 
(q0) edge[bend left, above] node{(} (q1)
(q1) edge[bend left, below] node{)} (q0)
(q1) edge[bend left, above] node{(} (q2)
(q2) edge[bend left, below] node{)} (q1)
(q2) edge[bend left, above] node{(} (q3)
(q3) edge[bend left, below] node{)} (q2)
(q3) edge[bend left, above] node{(} (q4)
(q4) edge[bend left, below] node{)} (q3);
\end{tikzpicture}}
& 
\makecell[lc]{0 \# 0 ( 1 \# 1 ) 0 \# 1 ( 2 \# 2 ) 1 \# 2 ( 3 \# 3 ) 2 \# 3 ( 4 \# 4 ) 3 \\
\textcolor{olive}{p \# p ( s \# s ) p \# s ( e \# e ) s \# e ( r \# r ) e <sep> \# r ( <ns> \# <ns> ) r <eos>} \\
\textcolor{olive}{s \# s ( t \# t ) s \# t ( f \# f ) t \# f ( e \# e ) f <sep> \# e ( <ns> \# <ns> ) e <eos>} \\
\textcolor{olive}{a \# a ( r \# r ) a \# r ( v \# v ) r \# v ( n \# n ) v <sep> \# n ( <ns> \# <ns> ) n <eos>} \\
}\\

\bottomrule
\end{tabular}
\vspace{6pt}
\caption{$\hat{h}^*_{k}$'s are RNNs trained to recognize $dyck_{1-k}$. We extract FSAs from RNN weights in light of their theoretical correspondence, and encode each FSA as a symbolic sequence. Such re-representation of $\hat{h}^*_{k}$'s makes it possible to learn model successors as decoder-only language models.}
\label{tab:FSAs}
\end{table}

\section{Schematic Diagrams}
\label{app:diagrams}

This section is intended to walk the reader through the definitions of various learning paradigms. We use schematic representations to aid the interpretation of their core differences. We also discuss the benefits and caveats of utilizing our schematics to reason about learning paradigms. 

\begin{figure}[h]
\centering
  \includegraphics[width=0.8\textwidth]{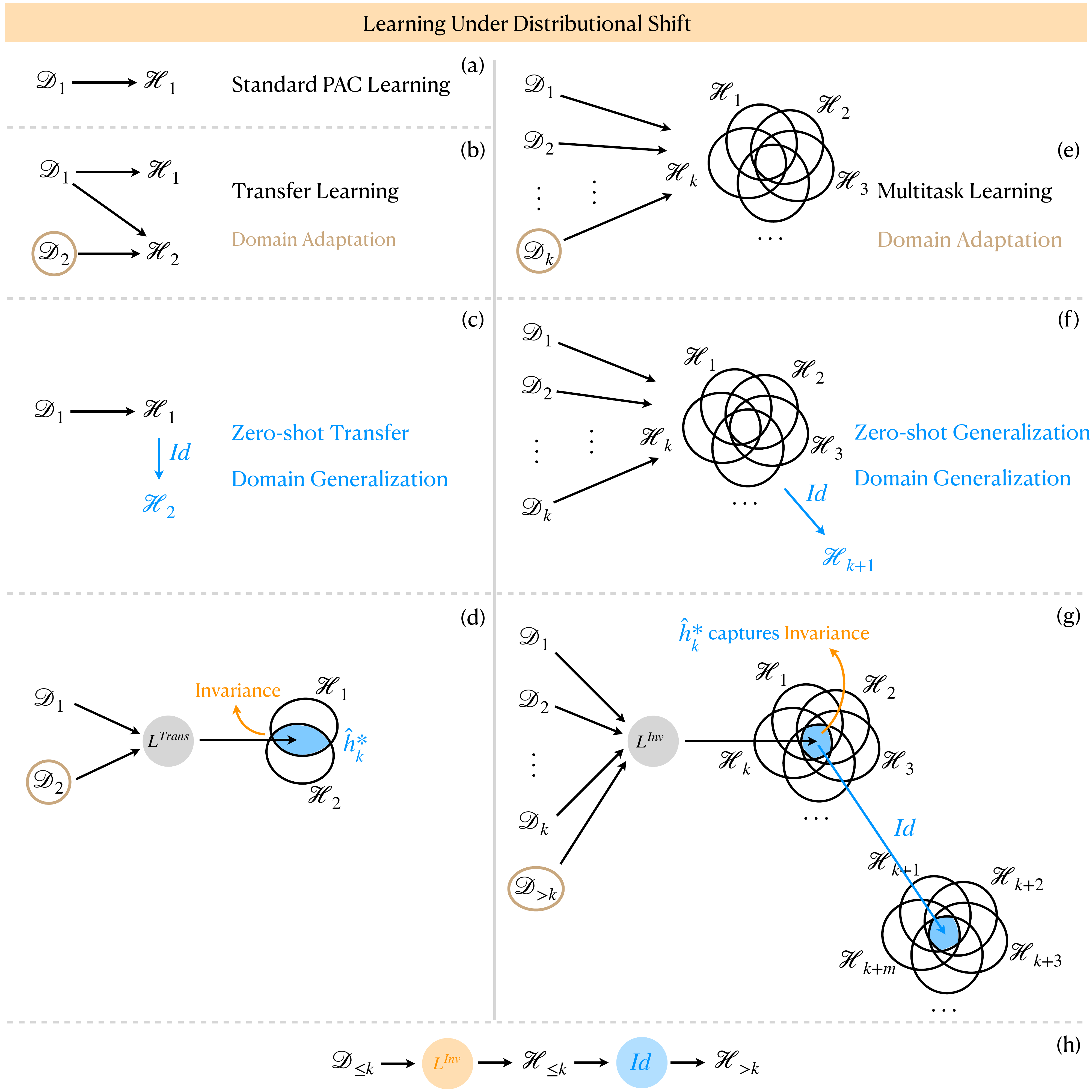}%
  \caption{We use a holistic term --- learning under distributional shift (\invlearner) --- to capture the focus on invariance and the static nature of the optimal hypothesis. \textbf{a.} In-domain PAC-learning is the most basic type of learning. \textbf{b-g.} Sub-frameworks encompassed by ``learning under distributional shift". \textbf{h.} A compact and unified diagram for ``learning under distributional shift".}
  \label{fig:a_myriad_of_learning}
\end{figure}

\begin{figure}[h]
\centering
  \includegraphics[width=1.0\textwidth]{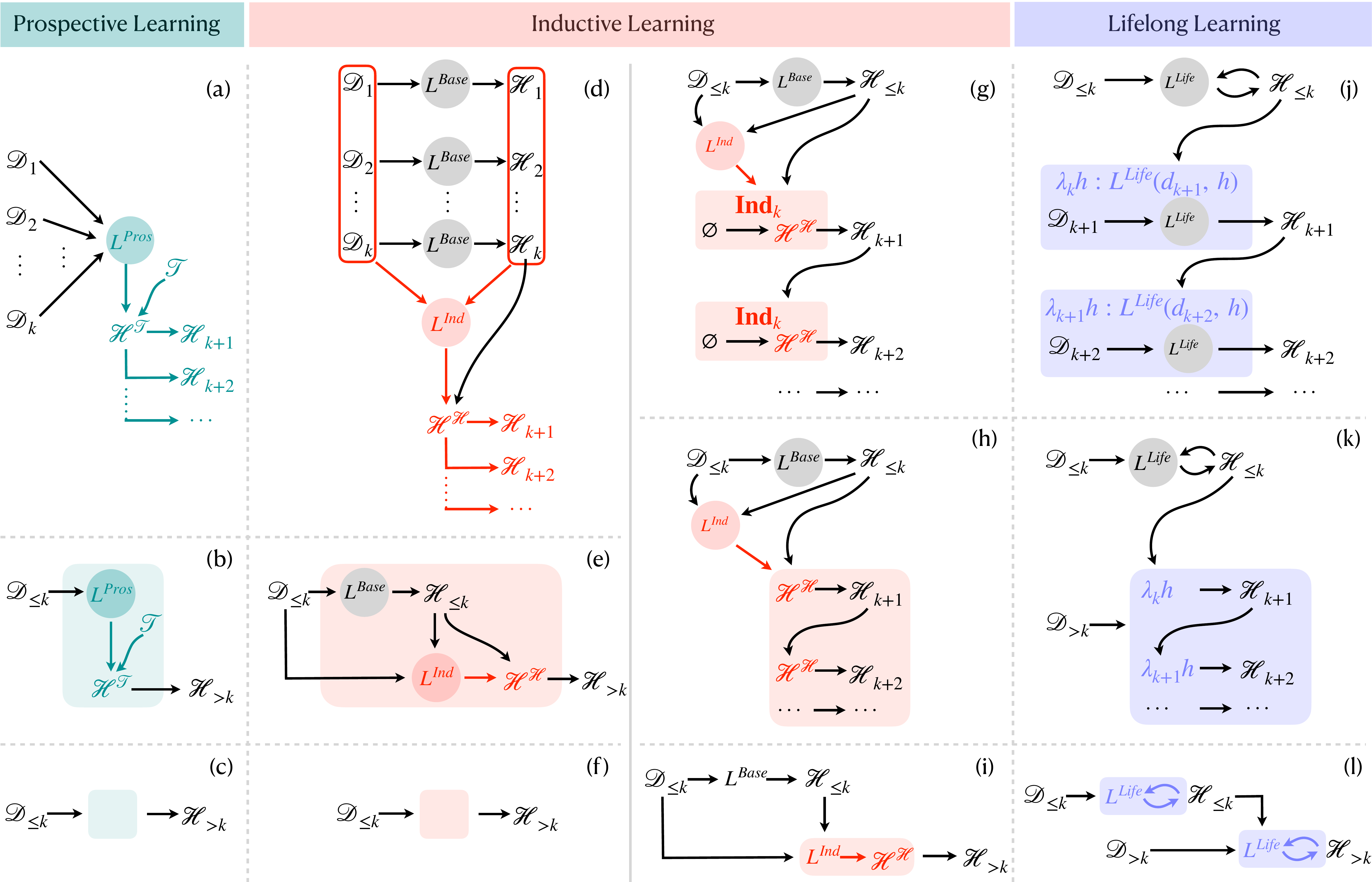}%
  \caption{\textbf{a \& d.} Standard diagrams for prospective (PL) and inductive learning (IL). \textbf{b \& e.} Demonstration of how syntactically transforming the graph reveals functionally equivalent components between PL and IL. \textbf{c \& f.} PL and IL can be reduced to the same abstract form --- inferring ``future" optimal hypotheses from observations encountered within finite horizon. \textbf{g \& j.} Syntactic manipulation of graphical elements also results in functional equivalence between LL and IL. Specifically, \Indk is \textit{functionally equivalent} to a learning algorithm instantiated as $\lambda_k h$. \textbf{h \& k.} \Indk$(\hat{h}^*_k)$ is functionally equivalent to \lifelonglearner$(d_{k+1}, \hat{h}^*_k)$, but eschews the need for future data beyond a finite $k$. \textbf{i \& l.} LL and IL are not identical despite maximal abstraction because LL constantly consumes new data.}
  \vspace{-6pt}
  \label{fig:pros_ind_life}
\end{figure}

The organization of learning frameworks is inherited from \cite{manyPAC}. We extend their organization to incorporate prospective learning (PL, Fig~\ref{fig:pros_ind_life}a) \cite{prospective, prospective_new} and inductive learning (IL, Fig~\ref{fig:pros_ind_life}d), and create diagrams for better illustration. 
The most basic learning framework is the in-distribution PAC learning (Fig~\ref{fig:a_myriad_of_learning}a) \cite{complexity_measures_for_indgen, ML-theory, learnability-stability-uniformcvg, Vapnik_statsLT}. 
Beyond the basic level, all types of learning involve the notion of OOD. Transfer learning (Fig~\ref{fig:a_myriad_of_learning}b) \cite{bengio2012, transfer_lowdim_representation, unlabeled_transfer, how_transferrable_Yosinski} makes use of experience in one domain to learn in another domain. Multitask learning (Fig~\ref{fig:a_myriad_of_learning}e) \cite{baxter2000indbias, ben-david_multitask_relatedness, caruana1997multitask, Chen_multitask, bayesian_multitask_UAI, Kumar_multitask, Lee_multitask} straightforwardly expands from two to many domains. Domain adaptation is subordinate to transfer and multitask learning, in which low-quality or unlabeled data from the target domain are provided to ease transfer.

Zero-shot transfer(generalization) is equivalent to transfer(multitask) learning with zero information about the target domain (Fig~\ref{fig:a_myriad_of_learning}[c,f]). The equivalence is in the sense that the optimal hypothesis obtained from the source domain(s) is mapped to the optimal hypothesis for the target domain via an identity function, $\mathbf{Id}$. Domain generalization can be a synonym for these scenarios.

In all the cases mentioned so far, the key to generalization is the capture of invariance by the in-domain optimal hypotheses. This assumes the existence of invariance, which translates to a non-trivial intersection of feasible hypothesis spaces (Fig~\ref{fig:a_myriad_of_learning}[d,g]) with respect to multiple domains. We use a holistic term --- learning under distributional shift (\invlearner) --- to capture the shared requirement for a non-evolving invariance-capturing hypothesis. This learning paradigm encompasses transfer/multitask learning, domain adaptation/generalization, and zero-shot transfer/generalization. A compact diagram unifying all subcases of \invlearner is shown in Fig~\ref{fig:a_myriad_of_learning}h.

\begin{figure}[h]
\centering
  \includegraphics[width=0.7\textwidth]{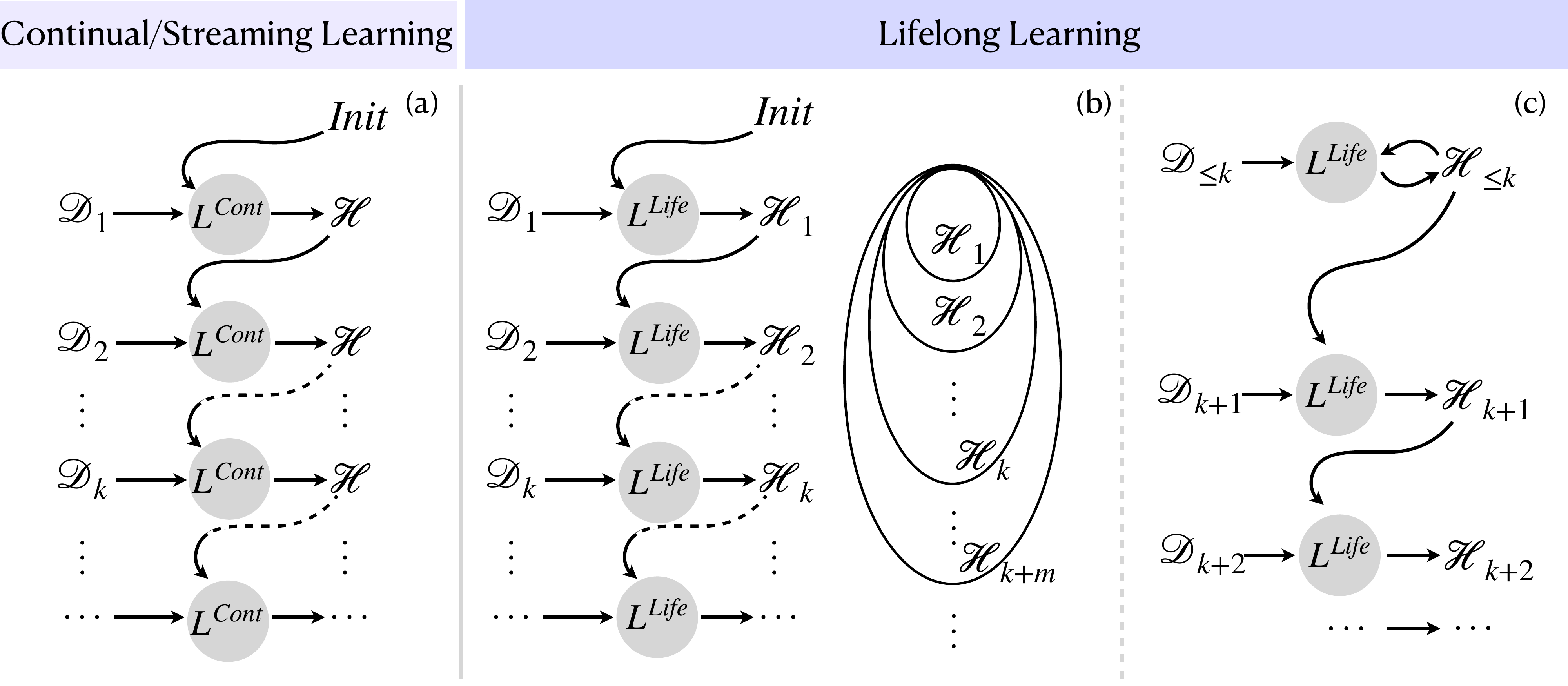}%
  \caption{Schematic illustration of streaming, continual and lifelong learning, all featuring a progressive manner of receiving data and inferring optimal hypotheses. \textbf{a.} New data arrive in individual examples and in batches for streaming and continual learning, respectively, which is a minor aspect that we do not distinguish in the diagrams. \textbf{b.} Lifelong learning extends continual learning by additionally requiring an explicit expansion of $\cH^{\mathrm{Ex}}$. \textbf{c.} The previous $k$ cycles in lifelong learning and be folded to separate the future from the past.}
  \vspace{-6pt}
  \label{fig:cont_life}
\end{figure}

Allowing for evolving the optimal hypothesis along with ongoing influx of data leads to continual learning (Fig~\ref{fig:cont_life} a) \cite{ring1994continual, survey_continualL_nlp, feature_extraction_formalization_continualL}. \citet{manyPAC} distinguishes streaming learning from continual learning in terms of whether new data arrive in individual examples or in batches, which we regard as minor and do not distinguish. Lifelong learning (LL, Fig~\ref{fig:cont_life} b) \cite{lifelongML, lifelong_robot, lifelong_supervisedL, continual_lifelong_review, lifelongLLM, EBMM_lifelong, ELLA_efficient_lifelongL} is a direct extension of continual learning, with the additional requirement for an explicit expansion of $\cH^{\mathrm{Ex}}$. 
Due to the progressive nature of lifelong learning, we can ``fold" the previous $k$ cycles in the diagram to separate the future from the past (Fig~\ref{fig:cont_life} c). In contrast to LL, we do not require an explicit expansion of $\cH^{\mathrm{Ex}}$ as we define IL. Instead, we focus on $\cH^{\mathrm{Fe}}$ when reasoning about the interplay between data and model progressions. When the learner's inductive biases hold constant, both $\cD_k$ and $\cH^{\mathrm{Ex}}$ can affect $\cH^{\mathrm{Lr}}$. Thus, introducing $\cH^{\mathrm{Fe}}$ as a new concept abstracts away whether the data distribution or $\cH^{\mathrm{Ex}}$ plays a greater role in shaping $\cH^{\mathrm{Lr}}$.

It can be seen that diagrams are nice tools for illustrating the \textit{syntax} of learning paradigms. 
In fact, LL, PL and IL are equivalent up to syntactic transformations over their graphical elements. \textbf{(1) Transforming PL into IL}: We can regard difficulty levels as timesteps, translating $\cD^{\cT}, \cH^{\cT}$ to $\cD_{\leq k}, \cH_{\leq k}$, respectively. Recall that PL requires producing $\hat{h}^*_{> k}$ altogether as a function of $k$. The same functionality is achieved in IL, where \Indk explicitly models how each $\hat{h}^*_m$ ($m>k$) can be derived from $\hat{h}^*_k$. Analogously, \Indk and $\hat{h}^*_k$ together specify a ``difficulty-indexed" sequence of hypotheses, $\hat{h}^*_{>k}$. Hence, the colored boxes in Fig~\ref{fig:pros_ind_life}[b,e] are \textit{functionally equivalent}, and when their inner details are abstracted away, PL and IL can be reduced to the same basic form (Fig~\ref{fig:pros_ind_life}[c,f]). 
\textbf{(2) Transforming LL into IL}: Assuming a given $d_{k+1}$, we can perform a currying operation\footnote{
    In functional programming \cite{curry1958combinatory, schonfinkel1967logic, slonneger1995pl}, \python{g :: (a, b) $\rightarrow$ c} can be \textit{curried} from \python{f :: a $\rightarrow$ (b $\rightarrow$ c)}.
} on \lifelonglearner, resulting in a partial function $\lambda_k h:$ \lifelonglearner$(d_{k+1}, h), h \in \cH_k$. Since \Indk and $\lambda_k h$ both map from $\cH_k$ to $\cH_{k+1}$, \Indk is \textit{functionally equivalent} to a learning algorithm instantiated as $\lambda_k h$ (Fig~\ref{fig:pros_ind_life}[g,j]). In this vein, \indlearner corresponds to ``learning a learning algorithm" based on the history streams of datasets and/or optimal hypotheses. As such, \Indk$(\hat{h}^*_k)$ and \lifelonglearner$(d_{k+1}, \hat{h}^*_k)$ can be treated as functionally equivalent operations (Fig~\ref{fig:pros_ind_life}[h,k]). However, \Indk is unary while \lifelonglearner is binary, highlighting the advantage of IL as eschewing the need for future data by inferring $\mathbf{Ind}_k$. For this reason, LL and IL cannot be reduced to identical basic forms even after maximal abstraction. Fig~\ref{fig:pros_ind_life}[i,l] shows the most compact forms of IL and LL. Their distinctive characteristics are emphasized via colored boxes.

The fact that we can derive equivalence among LL, PL and IL by manipulating their syntax has two implications. On the one hand, it shows that this paper does not introduce a fundamentally new primitive concept to machine learning, although the term ``model successors" may sound unfamiliar. Rather, the proposed learning framework amounts to a new \textit{arrangement} using existing concepts, such as distributions, hypotheses and learners. This underscores the flexibility and unification enabled by our formal notation, which aligns discussions about bespoke approaches to a shared common ground. On the other hand, meaningful comparisons must reside in the ``semantics" underlying syntax. Each syntactic arrangement uniquely implies which functions must be explicitly instantiated vs. many others that only implicitly exist. For example, any number of gradient descent steps can be viewed as a successor over models, as they amount to transformations in the hypothesis space. However, such functional equivalence between gradient descent steps to a model successor is implicit and without post hoc interpretations, no special significance is attached to a random gradient descent trajectory. What functions are explicitly instantiated vary across learning paradigms. Usually, these differences are only surfaced at an appropriate abstraction level. For example, Fig~\ref{fig:pros_ind_life}[b,e] reveal the difference between PL and IL while Fig~\ref{fig:pros_ind_life}[c,f] do not. A transformation between syntactic arrangements essentially involves the exchange of assumptions. For example, in IL, the removal of dependency on $\cD_{>k}$ is contingent on the assumption that $\cD_{>k}$ deviates from $\cD_{\leq k}$ in principled ways, and that the principles are identifiable during learning on $\cD_{\leq k}$. Comparisons across learning paradigms merely via syntactic relations are vacuous unless the exchange of assumptions is elaborated. 

To summarize, there are three takeaways for comparing learning paradigms: 1) What requires explicit instantiation matters; 2) The level of abstraction matters; 3) Meaningful comparisons can be made through the lens of assumption exchange.

\section{Challenges with Formalizing Inductive Generalization for Continuous Data}
\label{app:continuous_space}
There is no shortage of generalization challenges concerned with a continuous input space.
For example, the computer vision community is interested in generalizing detection to unseen objects \cite{OWR_2015, learn_the_unk_OWR_survey, OSR_extreme_value_theory, OWR_continual_active} or unseen scenes \cite{lost_domaingen, manyfacesrobustness}. The challenge associated with how discrete categories can be carved out of a continuous space through learning has a substantial literature of its own, such as category learning \cite{categoricallearning, concepts_and_categorization, categorylearningsupportexemplats} or concept learning \cite{perceptionforobjectconcepts, visualturingtests, individuals-stuffs-realkinds, tenenbaum-thesis}. The magnitude of continuous variables, such as contrast, luminance, sharpness, viewpoint \cite{risk-extrpl, deeperbroaderartier} may also go out-of-domain. It is unclear how a continuous space can be quantized into denumerable intervals. An artifical segmentation of continuous values does not inform the data successorship across intervals. The scope and nature of these difficulties need to be better understood before incorporating continuous cases under the formalization of inductive generalization.

\section{OODG While Not Evolving The Optimal Hypothesis}
\label{sec:not_evolvingOH}

This section surveys two broad categories of literature that tackles OODG assuming a static optimal hypothesis. Their achievements and obstacles shed light on how inductive learning should progress.

\vspace{-6pt}

\subsection{Generalization by Capturing Invariance}
\label{subsubsec:inv-capturing}
Classically, establishing theoretical generalization bounds under distributional shifts is of central concern in the field of domain generalization (Tab~\ref{tab:ELG_Inv}). Provable OODG is usually approached by imposing assumptions on the data divergence and/or properties of the target function \cite{theory_of_domaingen, impossibility_domainadapt, Firststep-extrpl, WILDS}. Classic results have settled the case where the source and target distributions share support, implied by the bounded density ratio assumption \cite{Firststep-extrpl}. Under the shared support assumption, generalization can be achieved practically by imbuing invariance-capturing mechanisms \cite{invariant-riskmin, domaingen-invariant-feature, hardtovary, GroupDROreview, GroupDRO, the-nth-thing}.

However, as modern intelligent machines face increasingly challenging scenarios, the conventional assumption on shared support can easily be violated \cite{provable_compo_lengen}. Without further assumptions, neural networks that perfectly fit the training data tend to exhibit arbitrarily erroneous behaviors in the region with zero training support \cite{GOTU, globality}. For example,  in graph-based reasoning, certain subgraphs tend to have vanishingly low support without careful sampling strategies, leading to extrapolation failure \cite{fate_faith}. In probabilistic autoregressive modeling, since the training data is very unlikely to span the entire space of sequences, the desired completion to any out-of-support prefix is nonidentifiable \cite{beyond-statsgen}. Suitable inductive biases must exist to account for the desired ``inductive leap" \cite{utgoff_ML_of_indbias}.

Classic theories cannot capture extrapolation behaviors on input outside the training support. As such, several recent studies have strived to close this gap. We view our work as strengthening the foundations of these lines of inquiry. \citet{Firststep-extrpl} does not assume shared support but requires matching marginal distributions and non-degenerate covariates among feature coordinates. \citet{extrapolation-by-transduction} similarly assumes marginal coverage together with a restricted target function class. Inductive generalization could benefit from extending this line of investigation with support mismatch to (a) (infinitely) many domains with progressive shifts and (b) provable inductive learning conditions. Such conditions should account for the divergence between optimal hypotheses inferred by the base-learner, and the properties of the target function class for learning model successors.

\vspace{-6pt}

\subsection{Generalization by Inference Time Scaling (ITS)}
\label{subsubsec:ITS}

ITS allows for predictions on unseen problem sizes, which can be enabled by recurrent architectures \cite{Avideepthinking} or non-recurrent architectures equipped with autoregressive decoding \cite{metageneration}. In the former, two families of approaches are most relevant to inductive generalization problems, both having the goal of simulating a recursive algorithm: (1) \textit{Deep thinking systems}, featuring looping ResNet or Transformer blocks \cite{Avideepthinking, thinkforlonger, adaptive_recurrent_vision}, and (2) \textit{Neural programmers}, aiming to explicitly model the execution traces of Turing machines \cite{recursiveNPI_provable, UTforlengen, NTM, NPL, NPI}. Provable extrapolation to unseen numbers of recursive steps has been established based on the correct realization of each individual recursive step \cite{recursiveNPI_provable}. One limitation of these lines of work lies in that models themselves do not learn to decompose a problem into low-level algorithmic steps, which is precisely the nontrivial part of problem solving \cite{RRNN-RNRN, seq2seq_learnability}. Future work is likely to see how to learn the correct decomposition that admits recursive modeling.

The latter category for ITS --- non-recurrent architectures paired with autoregressive decoding --- has recently gained traction due to the unprecedented ``zero-shot" ability of autoregressive LLMs \cite{ZS_videoimitators, ZS_planners, ZS_reasoners, ZS_trajectorygenerators, ZS_patternmachines, ZS_timeseries, ZS_proposers, ZS_retrievers}.
An emerging line of research attempts to formalize ``autoregressive learnability", i.e., AR-learnability \cite{AR-learnability, n_rconsistency}. 
However, two issues prevent these theoretical advances from informing practical choices. First, adequate learning depends on the data (consisting of long chain-of-thought sequences) to do the heavy lifting \cite{seq2seq_learnability}, at the expense of high computational and sample complexity \cite{AR-learnability}. 
Second, the realization of specialized decoding procedures demands external control. It is crucial to adopt modulated decoding procedures for AR generation to resemble program execution traces. For example, \citet{globality} introduces an ``inductive scratchpad" decoding format which relies on a special masking scheme and position reindexing. \citet{AR_lag_system} studies AR models under the conditions that (a) they have restricted attention windows, and (b) they are allowed to emit a pair of tokens within a single decoding step. \citet{turingprogramsscratchpad} develops a stylized scratchpad method that allows the simulation of a Turing machine, including operations analogous to tape memory updates. \citet{n_rconsistency} demonstrates provable length generalization when the scratchpad formulation satisfies ``(n, r)-consistency". Such a formulation requires (a) position indicators, resembling a tape head pointer, (b) strategies for embedding a ``multi-line input", and (c) two-sided padding to ensure the alignment of salient components with the center of the context window. All of them are open questions to be addressed before we can make stylized decoding strategies compatible with scalable pretraining setups \cite{RFWPs, pushdown_layers}. 

One unresolved problem common to all ITS approaches is the halting decision. Existing models usually lack the ability to decide on their own the optimal timing to halt. Previous works have largely worked around this problem by a) reporting performance once the ground-truth decoding length is reached \cite{UTforlengen}, b) selecting the best performance/confidence within an artificial computation budget \cite{UTforlengen}, c) relying on the generation of EOS \cite{globality, rule-extrpl} or d) hand-crafted halting patterns \cite{n_rconsistency}. Integrating techniques based on adaptive computation time \cite{ACT_for_rnns, adaptive_recurrent_vision} and dynamic halting \cite{recursiveNPI_provable, NPI, UT,PonderNet_dynamic_halting, memo_dynamic_halting} with ITS should be an important future venue. Furthermore, an intricacy that calls for caution is that the halting decision may itself be subject to poor OODG, when the model's internal states render ``unseen inputs" for the halting module during extrapolation\footnote{
    For example, \citet{NPI} reported that Neural-Programmer Interpreters can length-generalize bubble sort from 20 to 60, beyond which the ``pointer" associated with the halting decision starts to make incorrect advancements. Relatedly, the ``\textit{\textbf{eos}}-problem", referring to the extrapolation error due to immature emission of \textit{\textbf{eos}}, has been raised in the language modeling literature \cite{simple_arithmetic, eos_decision_and_lenextrpl, locattn_for_extrpl}.
}.  

\citet{superpolynomial} suggests three paths to transcend the limit imposed by bounded computation per AR step: grow a) runtime, b) number of parameters, or c) parameter size \textit{superpolynomially} in input length. ITS aims for (a), while suffering from the challenges we just discussed. Pursuing (b) and (c) requires model successors because growing the number of parameters or parameter size at inference time means making changes to the optimal model without new influx of data.

\section{Historical Insights for Defining \indlearner (cont.)}
\label{app:roadmap_cont}

We review a general allied literature for inductive learning and explains how they can be repurposed.

\paragraph{Bayesian Model Averaging (BMA)} \cite{mackay1992bayesian, Bayesian_for_NNs} uncovers the source of rich training signals for $L^{\text{Ind}}$. BMA offers an elegant way to record multiple moments along the learning course of $L^{\text{Base}}$, yielding a handful of $\hat{h}^*_k$ that predict a high likelihood of data \cite{PAC-BMA, approx-bayesian-ensembling, BMA}. The classic advantages of BMA lies in alleviating double descent and explaining generalization from a probabilistic view \cite{BMA}. The appeal of BMA for designing \indlearner is that it may help escaping the simplicity bias via simultaneous tracking of multiple basins of attraction in the loss landscape of \baselearner. Recall that our previous argument for the failure mode of \invlearner is that the simplicity bias would drive learning towards simpler hypotheses unless there are strong incentives for overriding this tendency. The simplicity bias largely constrains what a learner can \textit{arrive} at, but it does not constrain what hypotheses can be \textit{encountered} over the course of learning. It is likely that moments over the learning trajectory can inform more about $\hat{h}^*_{k+1}$ than $\hat{h}^*_k$ could. A Bayesian model average maintains a bag of compelling hypotheses and some of them are not minimizing simplicity. This significantly enriches the clues that a progression of (compelling) models could offer. Therefore, we believe that the probabilistic view of neural network learning embraced by BMA may shed light on both a) theorizing learnability conditions, and b) operationalizing the curation of training signals for \indlearner.

\paragraph{Symbolic Metaprogram Search} \cite{symbolic_metaprogram_search} describes a rule-learning system which has concretely realized all steps in Tab~\ref{tab:design_indlearner}. In their context, $h$ is a symbolic program. A transformation from $h_1$ to $h_2$ is a \textit{metaprogram} that revise programs. They also proposed a \textit{meta program learner} (MPL) that performs search over programs and metaprograms. MLP approximates MAP inference in a Bayesian posterior over metaprograms \cite{goodman2008rational, yang2022one}. It is demonstrated that MPL can effectively infer list functions \cite{cropper2020learning, inferLISP1975} from input-output pairs. The appeal of MPL is that it provides representations for both members of $\cH$ and members of $\cH^{\cH}$, together with a full-fledged learning algorithm for navigating the space of metaprograms in search for an optimal one. The downside is that the strong symbolic flavor of MPL limits its practical viability. The symbolic nature was not a big concern when the original purpose of developing MPL was to explain human rule learning under restricted computation and data. However, it remains not yet clear how the connectionist counterparts to programs and meta-programs can be represented. We expect this to be the subject of future neurosymbolic studies.

\paragraph{Neural Architecture Search (NAS)} \cite{NASsurvey_elsken, efficient_NAS, bananas_bayesNAS, 1000NASpapers} is concerned with finding the best topology of neural networks in addition to the best parameter values. NAS is inspirational in terms of how the ``syntax" of $h$ can be compactly represented, for example an encoding of the hyperparameter profile, which may in turn suggest compact representations of a transformation on $h$. Specifically, if the syntax of $h$ is encoded into differentiable vectors \cite{DARTS_Yiming}, then transformations on $h$ can be straightforwardly deduced via vector arithmetics. While NAS informs about representations of elements in $\cH$, and perhaps $\cH^{\cH}$, how the \textit{optimal} element in $\cH^{\cH}$ can be learned remains outside the realm of NAS. NAS operates by applying transformations in $h$ until a reasonable $\hat{h}^*$ is found. Thus, the final output of NAS is still a hypothesis (equivalent to what our \baselearner would output) rather than an optimal mapping over hypotheses. Inductive generalization is more likely to benefit from a particular branch of NAS that adopts evolutionary algorithms to search over topologies \cite{NAS_evolutionary_hierarchical, evolutionaryNAS_survey}. For example, LEMONADE \cite{LEMONADE} maintains the entire pareto frontier of topologies, guiding the warm-starting of a child network from their trained parents. This can be thought of as learning an optimal transformation from $\hat{h}^*_k$ to $\hat{h}^{\text{init}}_{k+1}$ which specifies the best initial point for learning $\hat{h}^*_{k+1}$. However, additional optimization steps are required as well as data from $\cD_{k+1}$, which does not conform to our inductive learning setups. Upgrading the NAS+evolutionary algorithm to one that directly outputs $\hat{h}^*_{k+1}$ without further optimization would bring us closer to an inductive learner.

\paragraph{Curriculum Learning (CL)} has two branches \cite{Elman1993, curriculum_survey}: a ``model progression" branch where a curriculum is embodied by growing capacities of the learner, and a ``data progression" branch where a curriculum is induced by growing complexities of the data \cite{GOTU, bengio2009curriculum}. The model progression branch is more relevant to designing \indlearner. Early representatives of the model curriculum include the Cascade-Correlation architecture \cite{cascade-correlation} and Dynamic Node Creation networks \cite{DNC}. Both approaches simultaneously optimize network parameters and topology by starting from a single ``unit" and sequentially adding new units. The core arguments of curriculum learning is that the extra requirement of evolving network capacity is not an added burden, but a desired degree-of-freedom \cite{three_constructive_algorithms}, and that without evolving from a small capacity, learning could be retarded \cite{Elman1993}. Arguments for the importance of capacity growth are developed in parallel in cognitive science under the term ``shaping" \cite{shaping}. Therefore, CL has insights to offer regarding the representation of a transformation from $h_1$ to $h_2$ such that $h_2$ is guaranteed to have greater capacity. Such representations of $\cH^{\cH}$ are more useful than those considered by NAS because they explicitly embody a capacity growth. Future works should flesh out the alignment between the difficulty progression (\S\ref{sec:difficulty_progression}) underlying cascaded training experiences and capacity growth underlying \baselearner's outputs.

\paragraph{Adapters} have gained tremendous attention regarding the parameter-efficient finetuning of large language models (LLMs) \cite{peft_survey, efficient_llm_survey}. An adapter straightforwardly specifies the difference between two hypotheses, thereby specifying a transformation from one to another. An adapter is a compact representation thanks to their low-rank nature. It is possible to treat the application of \Indk to $\hat{h}^*_k$ as applying an adapter. Most works in the LLM finetuning literature train one adapter per finetuning task \cite{llm-adapters, adapterhub}. To move beyond one-time usage, existing work has proposed meta-tuning \cite{meta-tuning-using-hypernetwork, meta-tuning-with-NAS}, which refers to the process of finding the optimal meta-aspects of adapters applicable to a breadth of downstream adaptation scenarios. To repurpose adapters for inductive learning, the question is how an optimal adapter can be learned so that applying it recursively keeps yielding optimal models that handle progressively difficult tasks. It is potentially promising to expand the line of meta-tuning research with the aim of finding an adapter that correctly embodies capacity growth (\S\ref{sec:indlearner}).

\end{document}